\documentclass[runningheads]{llncs}

\usepackage{eccv}

\usepackage[table]{xcolor}

\usepackage{eccvabbrv}

\usepackage{graphicx}
\usepackage{booktabs}

\usepackage{booktabs}
\usepackage{multirow}
\usepackage[table]{xcolor}
\usepackage{tabularx, colortbl}
\usepackage[table]{xcolor}
\usepackage{hhline}
\usepackage{nicematrix}
\usepackage{graphicx}

\definecolor{lightblue}{RGB}{217,234,247}
\definecolor{lightcyan}{RGB}{221,243,250}
\definecolor{lightgreen}{RGB}{228,240,228}
\definecolor{ForestGreen}{RGB}{34,139,34}

\usepackage[accsupp]{axessibility}  
\usepackage{wrapfig}

\usepackage{hyperref}
\usepackage{float}
\usepackage{orcidlink}
\usepackage{placeins}

\usepackage{longtable}
\usepackage{makecell}
\usepackage{adjustbox}
\usepackage{changepage}
\usepackage{threeparttable}
\usepackage{subcaption}
\usepackage{algorithm2e}
\usepackage{algpseudocode}
\usepackage{algorithmicx}
\usepackage{algcompatible}
\usepackage{overpic}
\usepackage{enumitem}
\setlist[itemize,1]{label=$\bullet$}

\begin{document}

\title{HyFL-CLIP: Hyperbolic Fine-Tuning of CLIP for Robust Long-Context Understanding}
\titlerunning{Hyperbolic Fine-Tuning of CLIP for Robust Long-Context Understanding}
\author{
Ji Ha Jang$^{1,*}$\orcidlink{0009-0003-6310-8262},\qquad
Hayeon Kim$^{1,*}$\orcidlink{0009-0008-7461-9750},\qquad
Chulwon Lee$^{2}$\orcidlink{0009-0003-4668-3644},\qquad \\
Junghun James Kim$^{2}$\orcidlink{0009-0001-8327-9662},\qquad  
Se Young Chun$^{1,2,3,\dagger}$\orcidlink{0000-0001-8739-8960}}

\authorrunning{JH Jang, H Kim \textit{et al.}}

\institute{$^1$Dept. of Electrical and Computer Engineering, $^2$IPAI, $^3$INMC \& AIIS, \\
Seoul National University, Republic of Korea \\ 
\email{\{jeeit17, khy5630, chul0e, jonghean12, sychun\}@snu.ac.kr}}
\maketitle
\def\thefootnote{*}\footnotetext{Authors contributed equally. $^{\dagger}$ Corresponding author.}

\vspace{-0.7cm}
\begin{abstract}
CLIP (Contrastive Language-Image Pre-training) has become a de facto paradigm for image-text alignment, but it struggles with long-context descriptions ($>77$ tokens) due to absolute positional encoding and pretraining on short captions. In long contexts, sentences are often reordered, summarized, or partially omitted. Although prior works extend CLIP with longer positional encodings, they often suffer from degraded image-text alignment under such text perturbations. We attribute this limitation to the Euclidean contrastive objective, which enforces strict one-to-one matching and lacks explicit mechanisms for modeling hierarchical relationships between global context and its constituent elements. To address this issue, we propose HyFL-CLIP, a hyperbolic fine-tuning framework that distills the well-established text-image alignment learned in Euclidean CLIP into hyperbolic space via cross-manifold similarity distillation, leveraging its geometry to capture hierarchical and entailment relations. Our method models hierarchical semantics by linking summarized token-wise features, long-context descriptions, constituent short textual components, and images, capturing part–whole relationships via hyperbolic entailment with Einstein midpoint aggregation. Experiments on diverse benchmarks, including long-context cross-modal retrieval, cross-modal retrieval with caption perturbations, intra-modality retrieval, and short-text cross-modal retrieval, show that HyFL-CLIP achieves more robust long-context understanding. In particular, it yields up to 19.5\% improvement in long-text cross-modal retrieval under textual perturbations over the best prior method. We also show HyFL-CLIP can be seamlessly integrated into other model frameworks by applying it to Stable Diffusion XL (SDXL). The project page is available at \url{https://janeyeon.github.io/hyflclip}.
\end{abstract}

\section{Introduction}

\begin{figure}[tb]
  \includegraphics[width=\linewidth]{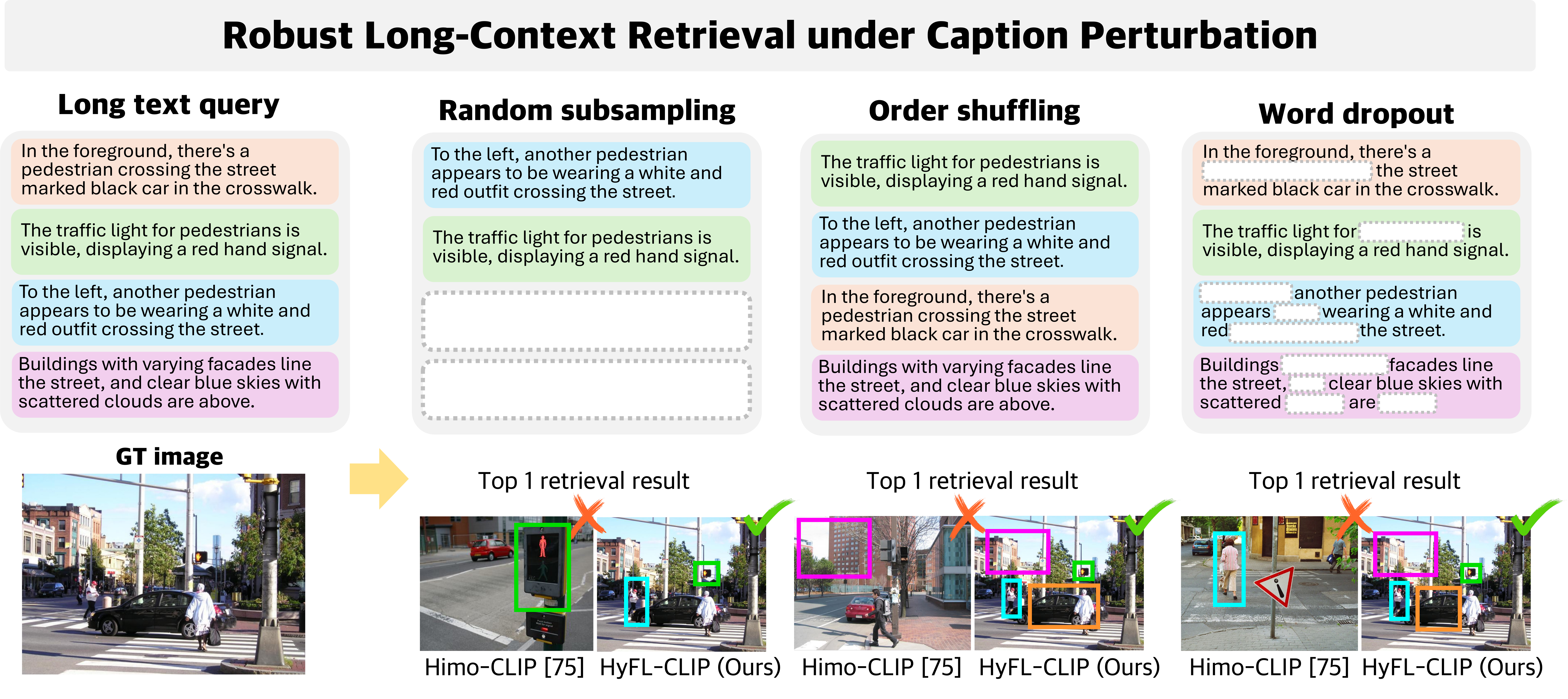}
  \caption{\textbf{Robust long-context retrieval under caption perturbation.} In long contexts, sentence order or structure may change preserving overall semantics, making robust text-image alignment essential. However, prior works on long-context understanding CLIP often exhibit degraded alignment under such perturbations. For example, even when only the sentence order is shuffled (Order shuffling), they fail to retrieve the correct image, missing key semantic elements such as the red pedestrian traffic signal, the black car, and the pedestrian wearing a white and red outfit.}
  \label{fig:motivation}
\end{figure}

Vision-language contrastive pre-training has laid the foundation for a wide range of vision-language learning tasks. Models such as CLIP~\cite{Radford2021LearningTV}, ALIGN~\cite{jia2021scaling} have achieved remarkable performance in text-image alignment by projecting image and text representations into a shared embedding space via contrastive learning. As a result, such models have been widely adopted not only for downstream tasks such as zero-shot classification and retrieval, but also for broader applications including image generation~\cite{rombach2022high, podell2023sdxl} and large multi-modal systems~\cite{liu2023visual, zhu2023minigpt, li2023blip, alayrac2022flamingo}. 

However, long-text inputs ($>77$ tokens) remain challenging for CLIP, as it is primarily trained on datasets composed of short captions and inherently supports a maximum input length of $77$ tokens~\cite{zhang2024long}. Existing approaches typically address this by extending positional encoding and aligning coarse and fine grained features in CLIP~\cite{wu2025himo, zhang2024long, najdenkoska2024tulip, asokan2025finelip}, and designing training strategies to better learn long-text representations~\cite{wang2025fix, feng2025retaining, najdenkoska2024tulip}. Although prior works on CLIP finetuning for long-context understanding~\cite{zhang2024long, wu2025himo, najdenkoska2024tulip, asokan2025finelip, wang2025fix, feng2025retaining} have significantly improved long text-image retrieval performance, we observe that they remain highly sensitive to (i) changes in the number or ordering of sentences that constitute the long context and (ii) the deletion or substitution of only a few words (See Fig.~\ref{fig:motivation}). This limitation may partly stem from CLIP's training paradigm, which primarily relies on one-to-one text-image matching and does not explicitly model entailment or part-whole relationships among semantic components. As a result, the global meaning of a long description can become concentrated on a few salient tokens, making the model sensitive to the removal or reordering of tokens or sentences and leading to degraded image–text alignment. Modeling long descriptions therefore requires capturing hierarchical relationships between the global context and its constituent semantic components.

Recently, hyperbolic space has been actively explored in vision–language contrastive learning to better model hierarchical relationships~\cite{desai2023hyperbolic, ramasinghe2024accept, pal2024compositional}. Unlike Euclidean space used by CLIP, hyperbolic space has negative curvature and naturally represents tree-like structures. However, most hyperbolic Vision-Language Models (VLMs)~\cite{desai2023hyperbolic, ramasinghe2024accept, pal2024compositional, yang2024hyperbolic, srivastava2025hypervlm, he2025hypercore} or Large Language Models (LLMs)~\cite{he2025helm} are trained from scratch, limiting their ability to leverage strong pretrained Euclidean models such as CLIP~\cite{Radford2021LearningTV}. While recent studies have begun exploring fine-tuning approaches, they typically focus on single-modality adaptation or specific tasks~\cite{yang2024hyperbolic, peng2025understanding, zhao2025fine, zhang2026parameter, wang2025learning, li2025text}.

We propose HyFL-CLIP (Hyperbolic Fine-tuning for Long-context CLIP), which enhances robust long-context understanding by modeling semantic part-whole relationships. HyFL-CLIP transfers the well-established text-image alignment learned by Euclidean CLIP into hyperbolic space. To the best of our knowledge, this is the first work to leverage Euclidean vision–language relationships for generalized tasks in hyperbolic space. We summarize token-wise features using Einstein midpoint aggregation to connect long-context descriptions and their constituent elements with the visual modality. Part-whole relationships are modeled through hierarchical entailment, which relaxes strict one-to-one matching imposed by contrastive learning and aligns semantically related elements to better capture hierarchical structures. We demonstrate that HyFL-CLIP outperforms prior arts~\cite{wu2025himo, wang2025fix, feng2025retaining, zhang2024long, najdenkoska2024tulip, asokan2025finelip} across diverse long-context benchmarks. By effectively leveraging hyperbolic space, our model remains robust even when parts of the input text are missing or perturbed, whereas Euclidean baselines exhibit larger performance drops. Furthermore, HyFL-CLIP achieves up to 19.5\% performance improvement on long-context cross-modal retrieval under caption perturbations, highlighting its ability to more robustly capture hierarchical semantic relationships in long contexts. We further adapt HyFL-CLIP to SDXL~\cite{podell2023sdxl}, demonstrating that it can be seamlessly integrated into existing frameworks. Our contributions are as follows:
\begin{itemize}
\item We introduce HyFL-CLIP, a hyperbolic fine-tuning framework that transfers the well-established image–text alignment learned by Euclidean CLIP into hyperbolic space, enabling well-pretrained Euclidean Vision–Language Models to operate effectively in hyperbolic representations.
\item We bridge Euclidean and hyperbolic representations via cross-manifold similarity distillation, while aligning image and text using a hyperbolic geodesic contrastive loss. A hierarchical entailment loss further relaxes strict one-to-one contrastive matching, enabling the model to better capture relationships between global context and its semantic components.
\item HyFL-CLIP achieves strong performance across diverse long-context understanding benchmarks, consistently outperforming Euclidean baselines. In particular, it shows improved robustness under caption perturbations, maintaining stable retrieval performance even when the input text is reordered, partially removed, or corrupted.
\end{itemize}

\section{Related Works}
\vspace{-0.2cm}
\subsection{Vision-language foundation models} 

Vision–Language Models (VLMs) learn a shared embedding space between images and text, becoming key components for cross-modal understanding~\cite{ge2023improving, pratt2023does, sain2023clip, jin2023refclip, sarkar2025crossover}. CLIP~\cite{Radford2021LearningTV} aligns image and text using a contrastive objective, achieving strong zero-shot performance in tasks such as image–text retrieval and classification. Building on this paradigm, many pretrained VLMs~\cite{hayeon-2024-iclr, lu2019vilbert, huang2020pixel, he2017fine, kiros2014unifying} have been developed and widely used for vision–language tasks.

Although contrastive loss of CLIP is powerful, it also has several limitations. First, contrastive learning treats all non-paired samples as negatives in binary~\cite{byun2024mafa}, even though some may share semantic overlap and act as false negatives in practice~\cite{huynh2022boosting, chen2021incremental}. In addition, CLIP-style training often relies heavily on the CLS token for global alignment, which can limit the model’s ability to capture hierarchical~\cite{geng2023hiclip, lv2025msg} or fine-grained structural associations~\cite{asokan2025finelip,zhong2022regionclip}.
To address these limitations, various approaches in Euclidean space have been explored~\cite{huynh2022boosting,chen2021incremental, geng2023hiclip, asokan2025finelip, zhong2022regionclip} to improve the contrastive learning framework. In this work, we instead transfer the well-trained similarity geometry of CLIP into hyperbolic space and further refine it using hyperbolic entailment. This allows the model to more robustly retrieve correct image-text pairs even when long-context descriptions are perturbed. 

\subsection{Long-context understanding with CLIP}
CLIP~\cite{Radford2021LearningTV} can face challenges when handling text sequences longer than $77$ tokens, as it is trained primarily on short captions and uses absolute positional encoding~\cite{zhang2024long}. A common strategy is to extend CLIP to longer sequences using interpolated positional encoding and coarse to fine cross-modal alignment strategies, as explored in Long-CLIP~\cite{zhang2024long} and HiMo-CLIP~\cite{wu2025himo}. Subsequent works further enhance token-wise fine-grained visual–textual correspondence~\cite{asokan2025finelip}. Other approaches introduce architectural modifications, such as relative positional encoding~\cite{najdenkoska2024tulip}, dual-branch training for jointly handling short and long contexts~\cite{wang2025fix}, and dual-teacher distillation frameworks for long-context learning~\cite{feng2025retaining}.

However, these approaches still rely on Euclidean embeddings with one-to-one image–text alignment and do not explicitly model hierarchical inclusion relationships, which can lead to performance degradation under semantically preserving text modifications. In contrast, we leverage hyperbolic space to explicitly model hierarchical structures that connect summarized token-wise features, long-context descriptions, and the short textual components that compose them. By linking global semantic summaries with their constituent textual elements, the model can robustly preserve semantic understanding even when parts of the text are removed, reordered, or perturbed.

\subsection{Hyperbolic representation learning in Vision-language models}
Hyperbolic geometry provides an embedding space  well suited for modeling fine-grained and hierarchical relationships~\cite{ganea2018hyperbolic, sala2018representation}. Due to its inherent tree-like structure, hyperbolic space naturally captures hierarchical data. As a result, hyperbolic embeddings have been widely applied in various domains, including graph, image understanding, and text~\cite{tifrea2018poincar, dhingra2018embedding, le2019inferring, liu2019hyperbolic, chami2019hyperbolic, sinha2024learning, khrulkov2020hyperbolicembeddings, yan2023hyp, atigh2022hyperbolic}.

Recently, several works have explored the use of hyperbolic geometry in VLMs~\cite{desai2023hyperbolic, ramasinghe2024accept, pal2024compositional, yang2024hyperbolic, srivastava2025hypervlm, he2025hypercore}. These models align image and text embeddings using geodesic contrastive losses, while further model hierarchical structure within the same modality~\cite{desai2023hyperbolic, ramasinghe2024accept} or across modalities via part–to–whole relations~\cite{pal2024compositional}. Also, some Large Language Models (LLMs)~\cite{he2025helm} adopt hyperbolic manifold to better capture the hierarchical structure inherent in language. 

Although prior works show promising results, they are generally trained from scratch and remain disconnected from well-pretrained Euclidean models, limiting their ability to leverage strong representations from models such as CLIP. While recent attempts address this gap by fine-tuning pretrained CLIP, they are often restricted to single-modality settings~\cite{wang2025learning, yang2024hyperbolic, zhang2026parameter} or specific tasks and experimental setups~\cite{peng2025understanding, zhao2025fine, li2025text}, which may limit their applicability in more general settings. In contrast, we perform hyperbolic fine-tuning starting from a pretrained CLIP model. By guiding the model with an entailment objective, our approach preserves CLIP’s strong representations while enabling the text embedding space to capture relationships across different levels of semantic abstraction, leading to improved performance in long-context understanding. 

\section{Preliminaries}
\subsection{Hyperbolic geometry in the Lorentz model} Hyperbolic space is a Riemannian manifold with constant negative curvature $-\kappa$, where $\kappa \in \mathbb{R}^{+}$. In this work, we employ the Lorentz (Minkowski) model as the underlying geometric space for hyperbolic fine-tuning, by distilling representations from a pre-trained Open-CLIP~\cite{cherti2023reproducible} model into the Lorentz manifold.  

We consider the $(n+1)$-dimensional Minkowski space 
$\mathbb{R}^{n+1}$ equipped with the Lorentzian inner product defined as below:
\begin{equation}
\langle \mathbf{p}, \mathbf{q} \rangle_{\mathbb{L}}
=
- p_{\text{time}} q_{\text{time}}
+
\langle
\mathbf{p}_{\text{space}},
\mathbf{q}_{\text{space}}
\rangle,
\end{equation}
where $\langle \cdot, \cdot \rangle$ denotes the standard Euclidean inner product.

Under this metric, the $n$-dimensional Lorentz manifold 
$\mathbb{L}^{n}$ is defined as the upper sheet of the two-sheeted hyperboloid given by:
\begin{equation}
\mathbb{L}^n
=
\left\{
\mathbf{p} \in \mathbb{R}^{n+1}
\;\middle|\;
\langle \mathbf{p}, \mathbf{p} \rangle_{\mathbb{L}}
=
-\frac{1}{\kappa},
\; p_{\text{time}} > 0
\right\}.
\end{equation}

Each point $\mathbf{p} \in \mathbb{L}^{n}$ is represented as below:
\begin{equation}
\mathbf{p}
=
\left[
p_{\text{time}},
\mathbf{p}_{\text{space}}
\right],
\qquad
p_{\text{time}}
=
\sqrt{
\frac{1}{\kappa}
+
\|\mathbf{p}_{\text{space}}\|^{2}
}.
\end{equation}
where $\mathbf{p}_{\text{space}} \in \mathbb{R}^{n}$ denotes the spatial component, $||\cdot||$ denotes the Euclidean norm, and the expression for $p_{\text{time}}$ follows from the Lorentz manifold constraint.

The geodesic distance between two points $\mathbf{p, q} \in \mathbb{L}^{n}$ is given by:
\begin{equation}
d_{\mathbb{L}}(\mathbf{p}, \mathbf{q}) = \sqrt{1/\kappa} \, \cosh^{-1}\left(-\kappa \langle \mathbf{p}, \mathbf{q} \rangle_{\mathbb{L}}\right).
\end{equation}
The hyperbolic radius of an embedding $\mathbf{p}$ corresponds to its hyperbolic distance from the hyperboloid origin $\mathbf{o}$, measured by $d_{\mathbb{L}}(\mathbf{p}, \mathbf{o})$.

\subsection{Tangent spaces} For each point $\mathbf{z} \in \mathbb{L}^n$, the tangent space at $\mathbf{z}$ is defined as: 
\begin{equation}
T_{\mathbf{z}}\mathbb{L}^n = \left\{ \mathbf{v} \in \mathbb{R}^{n+1} : \langle \mathbf{z}, \mathbf{v} \rangle_{\mathbb{L}} = 0 \right\},
\end{equation}
which forms an $n$-dimensional Euclidean vector space consisting of all vectors orthogonal to $\mathbf{z}$ under the Lorentzian inner product.
A tangent vector $\mathbf{v} \in T_{\mathbf{z}}\mathbb{L}^n$ can be mapped back to the hyperboloid via the exponential map,
\begin{equation}
\exp^{\kappa}_{\mathbf{z}}(\mathbf{v}) =
\cosh(\sqrt{\kappa} \, \| \mathbf{v} \|_{\mathbb{L}})\mathbf{z} +
\frac{\sinh(\sqrt{\kappa} \, \| \mathbf{v} \|_{\mathbb{L}})}{\sqrt{\kappa} \, \| \mathbf{v} \|_{\mathbb{L}}}\mathbf{v}.
\end{equation}
In contrast, the logarithmic map transports $\mathbf{p} \in \mathbb{L}^n$ to the tangent space at $\mathbf{z}$ as:
\begin{equation}
\log^{\kappa}_{\mathbf{z}}(\mathbf{p}) =
\frac{\cosh^{-1}(-\kappa \langle \mathbf{z}, \mathbf{p} \rangle_{\mathbb{L}})}%
{\sqrt{(\kappa \langle \mathbf{z}, \mathbf{p} \rangle_{\mathbb{L}})^2 - 1}}
\; \mathrm{proj}_{\mathbf{z}}(\mathbf{p}),
\end{equation}
where $\mathrm{proj}_{\mathbf{z}}(\mathbf{p}) = \mathbf{p} + \kappa \, \langle \mathbf{z}, \mathbf{p} \rangle_{\mathbb{L}} \mathbf{z}$ denotes the projection of $\mathbf{p}$ onto the tangent space  $T_{\mathbf{z}}\mathbb{L}^n$. In our formulation, the reference point $\mathbf{z}$ is set to the hyperboloid origin $\mathbf{o} = [\sqrt{1/\kappa}, \mathbf{0}]$. At this point, the tangent space $T_{\mathbf{o}}\mathbb{L}^n$ reduces to an $n$-dimensional Euclidean space, as tangent vectors have zero time components and are fully parameterized by their spatial coordinates, consistent with standard practice in hyperbolic representation learning~\cite{desai2023hyperbolic, pal2024compositional, ramasinghe2024accept}. 

\subsection{Einstein midpoint} In hyperbolic space, the centroid of a set of points is computed via the Einstein midpoint. We first project hyperbolic points from the hyperboloid model $\mathbb{L}^n$ to the Klein model $\mathbb{K}^n$, where a weighted average admits a closed-form expression. The Lorentz-Klein projection and its inverse are given by: 
\begin{equation}
\mathrm{\Pi}_{\mathbb{L}\to\mathbb{K}}(\mathbf{p})
=
\frac{\mathbf{p}_{\mathrm{space}}}{p_{\mathrm{time}}}
=
\frac{\mathbf{p}_{\mathrm{space}}}{\sqrt{\frac{1}{\kappa}+\|\mathbf{p}_{\mathrm{space}}\|^2}}
,\qquad
\mathrm{\Pi}_{\mathbb{K}\to\mathbb{L}}(\mathbf{k})
=
\frac{(1,\mathbf{k})}{\sqrt{\kappa (1-\|\mathbf{k}\|^2)}},
\end{equation}
where $\mathrm{\Pi}_{\mathbb{L}\to\mathbb{K}}$ maps a point $\mathbf{p}\in\mathbb{L}^n$ to its Klein coordinate representation, and $\mathrm{\Pi}_{\mathbb{K}\to\mathbb{L}}$ denotes the inverse projection that lifts a Klein point $\mathbf{k}$ back onto the Lorentz hyperboloid.

Given a set of points $\{\mathbf{x}_j\}_{j=1}^N \subset \mathbb{L}^n$,
their Einstein midpoint $\bar{\mathbf{x}}$ is obtained by computing a Lorentz-factor-weighted mean in Klein coordinates as:
\begin{equation}
\bar{\mathbf{x}}
=
\mathrm{\Pi}_{\mathbb{K} \to \mathbb{L}}\left(
\frac{
\sum_{j=1}^{N}
\gamma_j \,\mathrm{\Pi}_{\mathbb{L} \to \mathbb{K}}(\mathbf{x}_j)
}{
\sum_{j=1}^{N} \gamma_j
}
\right),
\qquad
\gamma_j=
\frac{1}{
\sqrt{
1 -\kappa
\left\|
\mathrm{\Pi}_{\mathbb{L} \to \mathbb{K}}(\mathbf{x}_j)
\right\|^2
}
}.
\end{equation}

\section{Method}
In this section, we introduce our method HyFL-CLIP. We first describe the problem setting, and then present the key training objectives of our approach: short-text guided cross-manifold similarity distillation, hyperbolic geodesic contrastive learning, hierarchical entailment with Einstein midpoint aggregation, and an entropy regularizer for stabilizing hyperbolic embeddings. 

\subsection{Problem formulation}
HyFL-CLIP fine-tunes a pre-trained CLIP model, which uses separate encoders for images and texts to produce aligned representations in a shared embedding space. We denote the model as $f$, consisting of image encoder $f_v$ and a text encoder $f_t$. For a text-image pair $(I,T)$, the visual encoder produces a set of embeddings $f_v(I) \in \mathbb{R}^{(K+1)\times n}$, consisting of a class token 
$\tilde{\mathbf{v}}$ that captures global image representation and a set of token-wise embeddings $\{\tilde{\mathbf{v}}_{k}\}_{k=1}^{K}$ that encode local visual information. Similarly, the text encoder produces a set of embeddings $f_t(T) \in \mathbb{R}^{(L+1)\times n}$, consisting of a sentence-level token $\tilde{\mathbf{t}}$ that represents the overall semantic meaning of the text and token-wise embeddings $\{\tilde{\mathbf{t}}_{l}\}_{l=1}^{L}$ that encode token-level linguistic information.

In the original CLIP model, the input text length is limited to $77$ tokens due to the use of learned absolute positional embedding. Before being fed into $f_t$, text tokens are truncated to the first $77$ tokens. To enable CLIP models to handle longer contexts, prior works~\cite{wu2025himo, asokan2025finelip, zhang2024long, najdenkoska2024tulip, feng2025retaining, wang2025fix} extend the positional embedding via interpolation to support longer input sequences (\textit{e.g.}, up to $248$ tokens). 

Our goal is to improve long-context understanding by explicitly modeling hierarchical semantic relationships within long descriptions in hyperbolic space, while preserving the well-established short text-image alignment learned by CLIP. 

\subsection{Short-text guided cross-manifold similarity distillation}
In pre-trained CLIP, the similarity between short text–image pairs is well learned. Inspired by prior similarity-based distillation methods that transfer relational structure through pairwise similarity distributions~\cite{Tung_2019_ICCV, wu2023tinyclip}, we use this well-established geometry as a reference when transferring the learned embedding space of CLIP into hyperbolic space. Unlike prior methods that typically distill similarities within the same geometric space, we perform cross-manifold similarity distillation, which aligns the similarity distributions of the Euclidean teacher and the hyperbolic student. Formally, let $\tilde{\mathbf{v}}_i \in \mathbb{R}^n$ denote the image embedding, and let $\tilde{\mathbf{t}}_i^s, \tilde{\mathbf{t}}_i^l \in \mathbb{R}^n$ denote the embeddings of the short and long texts corresponding to the same image, obtained from a pre-trained CLIP model that serves as the Euclidean teacher. Since CLIP is originally trained on short text–image pairs, we perform the distillation using the short-text embeddings to preserve the well-learned similarity geometry. We project these Euclidean embeddings into hyperbolic space via the exponential map at the origin: $\mathbf{v}_i = \exp^{\kappa}_{\mathbf{o}}(\tilde{\mathbf{v}}_i), \mathbf{t}_i^s = \exp^{\kappa}_{\mathbf{o}}(\tilde{\mathbf{t}}_i^s), \mathbf{t}_i^l = \exp^{\kappa}_{\mathbf{o}}(\tilde{\mathbf{t}}_i^l)$, where $\mathbf{v}_i, \mathbf{t}_i^s, \mathbf{t}_i^l \in \mathbb{L}^n$.

We define the Euclidean teacher similarity $S^{\mathrm{E}}$  as the cosine similarity between the short-text and image embeddings. In hyperbolic space, the student similarity $S^{\mathrm{H}}$ is defined as the negative geodesic distance in the Lorentz model (Eq.~(4)). Formally, 
\begin{equation}
S^{\mathrm{E}}(\tilde{\mathbf{t}}_i^{s},\tilde{\mathbf{v}}_j)
=
\frac{
\langle
\tilde{\mathbf{t}}_i^{s},
\tilde{\mathbf{v}}_j
\rangle
}{
\|
\tilde{\mathbf{t}}_i^{s}
\|
\|
\tilde{\mathbf{v}}_j
\|
},
\qquad
S^{\mathrm{H}}(\mathbf{t}_i^{s},\mathbf{v}_j)
=
-\, d_{\mathbb{L}}\!\left(\mathbf{t}_i^{s}, \mathbf{v}_j\right).
\end{equation}

We construct probability distributions over image candidates induced by the similarity scores. Given a similarity matrix $S$ and temperature $\tau$, the distribution is defined as:
\begin{equation}
P(S,\tau)_{ij}
=
\frac{
\exp\!\left(S_{ij}/\tau\right)
}{
\sum_k
\exp\!\left(S_{ik}/\tau\right)
}.
\end{equation}
The Euclidean teacher distribution and hyperbolic student distribution are then obtained as $P^{\mathrm{E}} = P(S^{\mathrm{E}}(\tilde{\mathbf{t}}^{s},\tilde{\mathbf{v}}),\tau_E), P^{\mathrm{H}} = P(S^{\mathrm{H}}(\mathbf{t}^{s},\mathbf{v}),\tau_H)$. The cross-manifold distillation loss is defined as the Kullback-Leibler divergence between the teacher and student distributions, given by:
\begin{equation}
\mathcal{L}_{\text{distill}}
=
\frac{1}{B}
\sum_{i=1}^{B}
\mathrm{KL}
\left(
P^{\text{E}}_{i\cdot}
\;\|\;
P^{\text{H}}_{i\cdot}
\right),
\end{equation}
where $B$ denotes the batch size, $i$ indexes the query samples in the mini-batch, and $\mathrm{KL}(\cdot\|\cdot)$ denotes the Kullback-Leibler divergence.

\begin{figure}[tb]
  \centering
  \includegraphics[width=\linewidth]{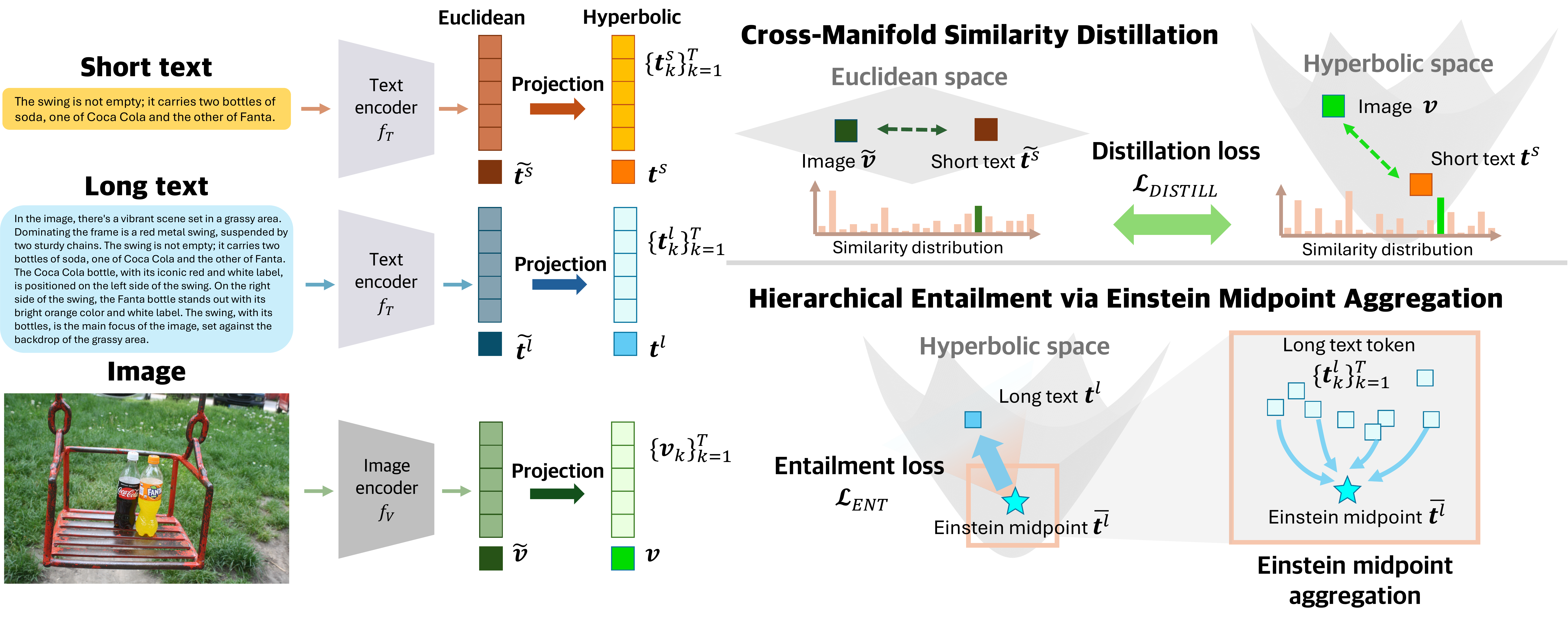}
  \caption{\textbf{Overview of our HyFL-CLIP framework.} Our HyFL-CLIP transfers Euclidean text–image alignment into hyperbolic space via a short-text guided cross-manifold similarity distillation. Hierarchical entailment with Einstein midpoint aggregation abstracts token-wise information within each modality and aligns it with a global representation. Hyperbolic geodesic contrastive loss aligns both long texts and their semantic components with the corresponding image, while an entropy regularizer stabilizes the embedding distribution.} 
  \label{fig:example}
\end{figure}
\vspace{-0.5cm}

\subsection{Hyperbolic geodesic contrastive loss}
After transferring Euclidean geometric relations through short text–image pairs, we further optimize the model using long text–image pairs through a geodesic contrastive objective on the Lorentz manifold $\mathbb{L}^n$. Following the hyperbolic InfoNCE formulation of~\cite{desai2023hyperbolic}, we define the image-text contrastive loss $\mathcal{L}_{\mathrm{itc}}$ in Lorentz space as follows:

\begin{equation}
L_{\text{info}}(\textbf{v}, \textbf{t}; \tau_c) = - \sum_i \log 
  \frac{
      \exp \left(-d_{\mathbb{L}}(\textbf{v}_i, \textbf{t}_i) / \tau_c \right)
  }{
      \sum_{\substack{k \neq i}} \exp \left(-d_{\mathbb{L}}(\textbf{v}_i, \textbf{t}_k) / \tau_c \right)
  },
\label{eq:contrastive_basic}
\end{equation} 
where $(\mathbf{v}_i,\mathbf{t}_k)$ denotes the matched text-image pair in the batch, while the remaining text embeddings $\{\mathbf{t}_k\}_{k \neq i}$ serve as negative samples. The temperature parameter $\tau_c$ controls the sharpness of the softmax distribution. 

We employ a bidirectional contrastive objective for both short text–image and long text–image pairs, defined as:
\begin{equation}
\mathcal{L}_{v \leftrightarrow t}^{s}
=
\frac{1}{2}
\left(
\mathcal{L}_{\mathrm{info}}(\mathbf{v}, \mathbf{t}^{s}; \tau_c)
+
\mathcal{L}_{\mathrm{info}}(\mathbf{t}^{s}, \mathbf{v}; \tau_c)
\right),
\end{equation}
\begin{equation}
\mathcal{L}_{v \leftrightarrow t}^{\ell}
=
\frac{1}{2}
\left(
\mathcal{L}_{\mathrm{info}}(\mathbf{v}, \mathbf{t}^{\ell}; \tau_c)
+
\mathcal{L}_{\mathrm{info}}(\mathbf{t}^{\ell}, \mathbf{v}; \tau_c)
\right).
\end{equation}

The final hyperbolic geodesic contrastive objective is given by:
\begin{equation}
\mathcal{L}_{\text{itc}}
= \mathcal{L}_{v \leftrightarrow t}^{\ell} +  \lambda_1 \mathcal{L}_{v \leftrightarrow t}^{s}, 
\end{equation}
where $\lambda_1$ is a hyperparameter.

\subsection{Hierarchical entailment via Einstein midpoint aggregation} Long-context understanding requires capturing a global representation as well as modeling the individual components and their semantic inclusion within the same modality. To capture such hierarchical structure, we construct a parent representation by aggregating token-wise features via a similarity-weighted Einstein midpoint. The weights are determined by their semantic similarity to the global representation from the opposite modality. We then enforce a hyperbolic entailment constraint between the aggregated representation and the corresponding global embedding. 

For clarity, we first describe the formulation for the long-text $\mathbf{t}^{\ell}$; the same procedure is applied to the image modality $\mathbf{v}$ . 
Let $\{\mathbf{t}^{\ell}_{i,k}\}_{k=1}^{K} \subset \mathbb{L}^n$ denote the long-text token embeddings of the $i$-th sample. 
To measure the semantic importance of each token-level feature, we compute attention weights $\alpha_{i,k}$ as follows:
\begin{equation}
\alpha_{i,k}
=
\frac{
\exp\!\left(- d_{\mathbb{L}}(\mathbf{t}^{\ell}_{i,k}, \mathbf{v}_i) / \tau_{\text{ent}}\right)
}{
\sum_{m}
\exp\!\left(- d_{\mathbb{L}}(\mathbf{t}^{\ell}_{i,m}, \mathbf{v}_i) / \tau_{\text{ent}}\right)
},
\qquad
\sum_{k} \alpha_{i,k} = 1.
\end{equation}
Then we abstract token-wise feature by calculating weighted Einstein midpoint, multiplying $\alpha_{i,k}$ to the Lorentz factor in Eq.~(9). The weighted Einstein midpoint is given as:
\begin{equation}
\bar{\mathbf{t}}^{\ell}_i
=\mathrm{\Pi}_{\mathbb{K} \rightarrow \mathbb{L}}(
\frac{
\sum_k
\alpha_{i,k}
\gamma_{i,k}
\mathrm{\Pi}_{\mathbb{L} \to \mathbb{K}}(\mathbf{t}^{\ell}_{i,k})
}{
\sum_k
\alpha_{i,k}
\gamma_{i,k}
}), \qquad
\bar{\mathbf{t}}^{\ell}_i \in \mathbb{L}^n.
\end{equation}

\begin{wrapfigure}{r}{0.43\linewidth}
\centering
\includegraphics[width=\linewidth]{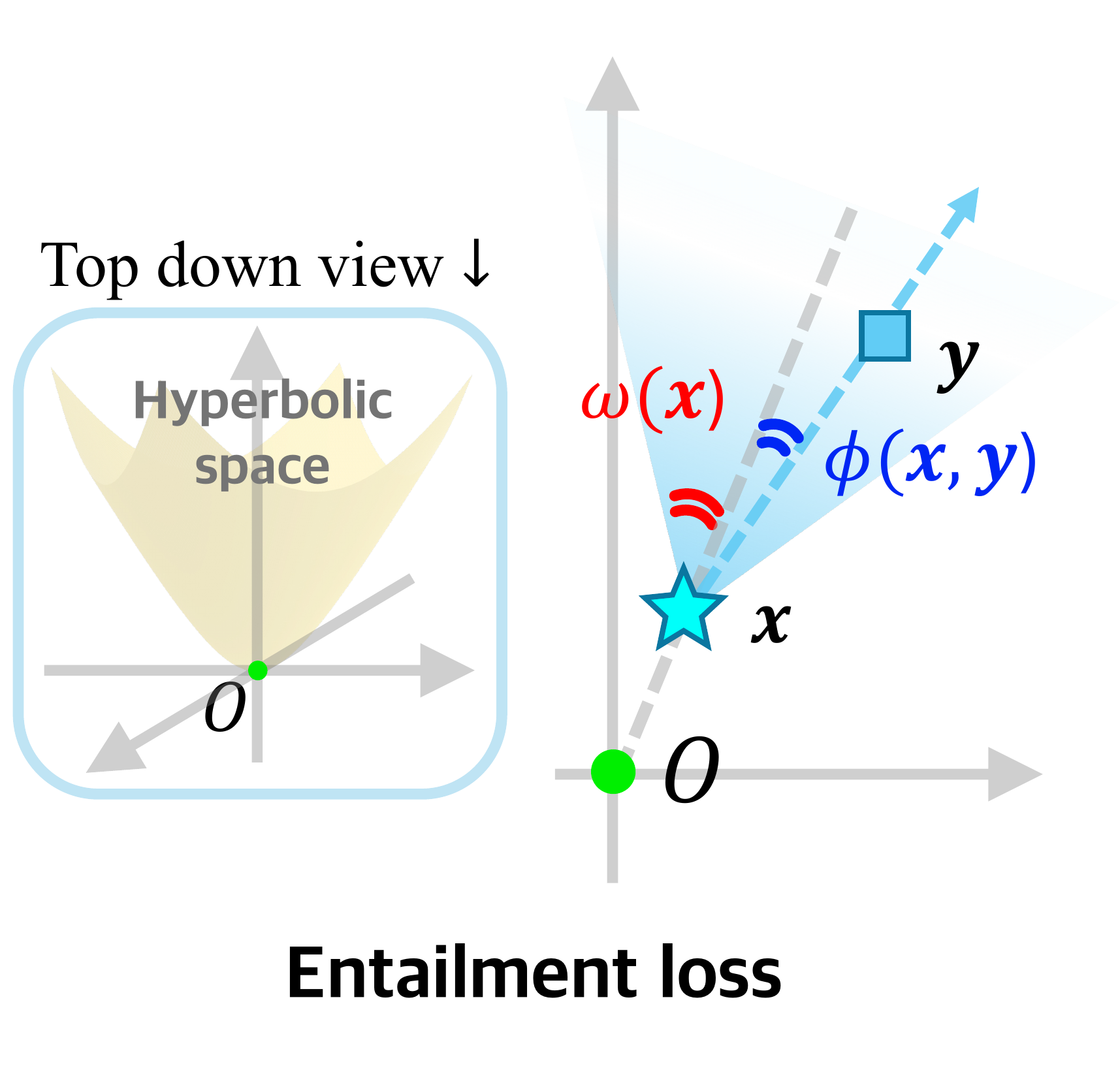}
\caption{\textbf{Entailment loss in hyperbolic space.} Adapted from the MERU~\cite{desai2023hyperbolic}, this figure illustrates the entailment loss in hyperbolic space. The $\phi(\mathbf{x}, \mathbf{y})$ measures the geodesic angle between the two embeddings, and $\omega(\mathbf{x})$ denotes the aperture of the entailment cone centered at $\mathbf{x}$. Geodesic angle between the two embeddings is used to determine if $\mathbf{y}$ lies within the entailment region of $\mathbf{x}$.}
  \label{fig:example}
\end{wrapfigure}

We interpret the aggregated representation $\bar{\mathbf{t}}_i^{\ell}$ as a more general semantic concept and enforce that the corresponding global embedding $\mathbf{t}_i^{\ell} \in \mathbb{L}^n$ lies within its entailment cone. Following the hyperbolic entailment formulation in~\cite{ganea2018hyperbolic, le2019inferring}, the half-aperture of the cone centered at $\bar{\mathbf{t}}_i^{\ell}$ is defined as:
\begin{equation}
\omega(\bar{\mathbf{t}}_i^{\ell})
=
\arcsin
\left(
\frac{2K}
{\sqrt{\kappa} \,
\|
\bar{\mathbf{t}}_i^{\ell}
\|_{\mathbb{L}}}
\right),
\label{eq:aperture}
\end{equation}
where $K$ is a constant controlling stability near the origin. 
We enforce the entailment constraint by penalizing violations of the cone boundary using the following loss (see Fig.~\ref{fig:example}):
\begin{equation}
\mathcal{L}_{\text{ent}}^{\mathbf{t}^{\ell}}
=
\frac{1}{B}
\sum_{i=1}^{B}
\max
\left(
0,
\phi(\bar{\mathbf{t}}_i^{\ell}, \mathbf{t}_i^{\ell})
-
\eta \, \omega(\bar{\mathbf{t}}_i^{\ell})
\right),
\end{equation}
where $\phi(\cdot,\cdot)$ denotes the hyperbolic angle between two embeddings in the Lorentz model. The above formulation is symmetrically applied to the image embeddings $\mathbf{v}$, yielding $\mathcal{L}_{\text{ent}}^{\mathbf{v}}$. The final hierarchical entailment loss is defined as:
\begin{equation}
\mathcal{L}_{\text{ent}}=\mathcal{L}_{\text{ent}}^{\mathbf{t}}
+\mathcal{L}_{\text{ent}}^{\mathbf{v}}.
\end{equation}

Thus, our final loss is given as:
\begin{equation}
\mathcal{L}
=
\lambda_2 \mathcal{L}_{\text{distill}}
+
\mathcal{L}_{\text{itc}}
+
\lambda_3 \mathcal{L}_{\text{ent}}
+
\lambda_4 \mathcal{L}_{\text{reg}}.
\end{equation}
Inspired by entropy-based regularization strategies used in prior works~\cite{kim2026uncertainty, grandvalet2004semi}, we introduce a radius-entropy regularization term:
\begin{equation}
\mathcal{L}_{\text{reg}}
=
- H(\mathbf{p}),
\qquad
p_i =
\frac{\exp\!\left(d_{\mathbb{L}}(\mathbf{t}_i^{\ell}, \mathbf{o})\right)}
{\sum_j \exp\!\left(d_{\mathbb{L}}(\mathbf{t}_j^{\ell}, \mathbf{o})\right)}.
\end{equation}
Here, $H(\mathbf{p}) = -\sum_i p_i \log p_i$ denotes the entropy of the hyperbolic radius distribution. This regularization promotes balanced utilization of hyperbolic radii. More details on the hyperparameter choices are in the supplementary material.

Intuitively, the entailment cone provides a geometric margin of tolerance. The global text embedding is trained to lie within the cone of the aggregated midpoint, which summarizes token-level features. When a perturbation removes or reorders tokens, the midpoint shifts, but the cone's nonzero aperture (Eq.~\ref{eq:aperture}) allows $\mathbf{t}_i^{\ell}$ to remain within the entailment region, preserving alignment. By contrast, Euclidean contrastive objectives enforce point-to-point matching with no such margin, making them sensitive to any shift in the aggregated representation.

\section{Experiments}
\subsection{Experimental setup}
\subsubsection{Training details.} Following Long-CLIP~\cite{zhang2024long}, we train HyFL-CLIP on dataset ShareGPT4V~\cite{chen2024sharegpt4v}, which contains $1.2$ million image–caption pairs with multi-sentence annotations and long captions averaging $143.6$ words. The batch size is $1024$ and we train our models for $2$ epochs. We set the learning rate to $1\times10^{-5}$ and the weight decay to $2.5\times10^{-2}$, and optimize the model using AdamW. Our model builds upon Open-CLIP, and we experiment with two CLIP architectures, CLIP-ViT-B/16 and CLIP-ViT-L/14.

\subsubsection{Evaluation setup.} We evaluate our model against other baselines~\cite{zhang2024long, najdenkoska2024tulip, wu2025himo, asokan2025finelip, feng2025retaining, wang2025fix, xie2025smartclip} on four different tasks: 

\noindent \textbf{(1) Zero-shot long/short caption cross modality retrieval.} We evaluate HyFL-CLIP on zero-shot long-caption text-image retrieval using DOCCI~\cite{onoe2024docci}, DCI~\cite{urbanek2024picture}, Long-DCI~\cite{najdenkoska2024tulip}, and Urban-1k~\cite{zhang2024long}, whose captions average $131.4$ to $174.2$ tokens. Performance was measured using Top-1 retrieval accuracy. To examine whether HyFL-CLIP successfully distills the text–image similarity geometry learned from short captions while adapting to long captions, we further evaluate short-text retrieval on COCO~\cite{lin2014microsoft} and Flickr30K~\cite{plummer2015flickr30k}.

\noindent \textbf{(2) Zero-shot long-caption cross-modal retrieval under caption perturbation.} Using all datasets from Task~(1), we perturb the captions to test whether models can robustly retrieve the correct images under semantic-preserving modifications. The perturbations include random word dropping ($p=0.5$), random subsampling of $n=2$ or $n=3$ sentences, sentence order shuffling, and removal of the first sentence. Each model is evaluated five times per perturbation, and Top-1 retrieval accuracy is averaged across all datasets. 

\noindent \textbf{(3) Text-to-text intra modality retrieval.} We perform text-to-text retrieval following the evaluation protocol of~\cite{mistretta2025cross}. For COCO~\cite{lin2014microsoft}, Flickr30K~\cite{plummer2015flickr30k}, and nocaps~\cite{agrawal2019nocaps} datasets, we ignore the images  and use the first caption of each image as the query, while the remaining captions of the same image are treated as positives. We use the Karpathy split~\cite{karpathy2015deep} for COCO and Flickr30K, and the validation split for nocaps. We further evaluate on purely textual datasets, including 20 Newsgroups~\cite{lang1995newsweeder} and IMDB Reviews~\cite{maas2011learning}. 

\noindent \textbf{(4) Text-to-image generation.} To qualitatively evaluate how HyFL-CLIP can be integrated into different model frameworks, we apply it to Stable Diffusion XL~\cite{podell2023sdxl} (SDXL). Since SDXL is originally trained with Euclidean text prompts, we encode text prompts from Long-DCI~\cite{najdenkoska2024tulip}, DrawBench~\cite{saharia2022photorealistic} and project them to the Euclidean tangent space using Eq.~(7) before image generation. We compare the results with Long-CLIP~\cite{zhang2024long} integrated with SDXL.

\subsection{Experimental results}

\subsubsection{Zero-shot long-caption cross-modal retrieval.} Tab.~\ref{tab:longcaptionretrieval} presents long caption cross-modal retrieval results. HyFL-CLIP (Ours) consistently outperforms Euclidean baselines across datasets and architectures. Comparisons with state-of-the-art hyperbolic VLMs~\cite{desai2023hyperbolic, pal2024compositional, kim2026uncertainty} are in the supplementary material.

\begin{table}[H]
\centering \caption{\textbf{Comparison of zero-shot long-caption cross-modal retrieval.} 
HyFL-CLIP (Ours) consistently outperforms existing long-context CLIP baselines across all datasets and model architectures. 
The best and second-best results are highlighted in \textbf{bold} and \underline{underline}, respectively. We report numbers from the original papers when available; otherwise, we evaluate using our own implementation (marked with $*$). 
Results marked with $\dagger$ are obtained using the checkpoints provided by the original authors.}
\label{tab:performance_comparison}

\setlength{\tabcolsep}{4pt}
\renewcommand{\arraystretch}{0.9}
\resizebox{\textwidth}{!}{ 
\begin{tabular}{l|lcccccccc}
\toprule
\multicolumn{2}{c}{\multirow{2}{*}{}} & \multicolumn{2}{c}{DOCCI~\cite{onoe2024docci}} & \multicolumn{2}{c}{DCI~\cite{urbanek2024picture}} & \multicolumn{2}{c}{Long-DCI~\cite{najdenkoska2024tulip}} & \multicolumn{2}{c}{Urban-1k~\cite{zhang2024long}} \\ \cmidrule(lr){3-4} \cmidrule(lr){5-6} \cmidrule(lr){7-8} \cmidrule(lr){9-10}
\multicolumn{2}{c}{} & I2T & T2I & I2T & T2I & I2T & T2I & I2T & T2I \\ \midrule
\multirow{8}{*}{\rotatebox{90}{ViT-B-16}} & Long-CLIP$^\dagger$~\cite{zhang2024long} & 63.10 & 71.49 & 59.88 & 61.28 & 42.21 & 48.38 & 79.40 & 79.60 \\
 & TULIP~\cite{najdenkoska2024tulip} & - & - & - & - & 50.20 & 50.60 & 88.10 & 86.60 \\
 & HiMo-CLIP*~\cite{wu2025himo} & 77.37 & \underline{79.35} & \underline{71.09} & \underline{69.93} & \underline{58.59} & \underline{57.00} & 89.20 & \underline{89.20} \\
 & FineLIP*~\cite{asokan2025finelip} & 77.16 & 79.14 & 69.38 & 68.03 & 57.18 & 55.22 & 89.30 & 86.90 \\
 & LongD-CLIP~\cite{feng2025retaining} & - & - & - & - & - & - & 87.20 & 87.30 \\
 & SmartCLIP~\cite{xie2025smartclip} & \underline{77.40} & 78.00 & 64.90 & 64.00 & 53.40 & 52.80 & \underline{90.00} & 87.40 \\
 & Fix-CLIP~\cite{wang2025fix} & - & - & 59.70 & 63.00 & - & - & 80.90 & 81.10 \\
 &  \cellcolor{gray!15}\textbf{HyFL-CLIP(Ours)} & \cellcolor{gray!15} \textbf{78.41} &  \cellcolor{gray!15}\textbf{81.12} &  \cellcolor{gray!15}\textbf{71.54} &  \cellcolor{gray!15}\textbf{71.79} &  \cellcolor{gray!15}\textbf{59.00} &  \cellcolor{gray!15}\textbf{58.75} &  \cellcolor{gray!15}\textbf{91.80} &  \cellcolor{gray!15}\textbf{91.10} \\ \midrule
\multirow{8}{*}{\rotatebox{90}{ViT-L-14}} & Long-CLIP$^\dagger$~\cite{zhang2024long} & 66.78 & 78.61 & 64.13 & 67.83 & 46.55 & 54.25 & 82.40 & 86.20 \\
 & TULIP~\cite{najdenkoska2024tulip} & 77.90 & 79.10 & - & - & 55.70 & 56.40 & 90.10 & 91.10 \\
 & HiMo-CLIP$^\dagger$~\cite{wu2025himo} & \textbf{82.35} & \underline{84.59} & \underline{74.59} & \underline{74.54} & \textbf{62.06} & \underline{61.94} & 93.00 & \underline{93.20} \\
 & FineLIP~\cite{asokan2025finelip} & \underline{82.20} & 83.10 & - & - & 60.80 & 60.70 & 93.20 & 93.00 \\
 & LongD-CLIP~\cite{feng2025retaining} & - & - & - & - & - & - & 91.90 & 90.80 \\
 & SmartCLIP~\cite{xie2025smartclip} & 81.60 & 82.50 & 68.20 & 69.80 & 57.60 & 58.50 & \underline{93.30} & 90.10 \\
 & Fix-CLIP~\cite{wang2025fix} & - & - & 65.10 & 66.70 & - & - & 86.80 & 87.70 \\
 &  \cellcolor{gray!15}\textbf{HyFL-CLIP(Ours)} &  \cellcolor{gray!15}82.12 &  \cellcolor{gray!15}\textbf{85.39} &  \cellcolor{gray!15}\textbf{74.74} &  \cellcolor{gray!15}\textbf{76.19} &  \cellcolor{gray!15}\underline{61.92} &  \cellcolor{gray!15}\textbf{63.93} &  \cellcolor{gray!15}\textbf{94.60} &  \cellcolor{gray!15}\textbf{94.30} \\ \bottomrule
\end{tabular}
}
\label{tab:longcaptionretrieval}
\end{table}

\subsubsection{Zero-shot long-caption cross-modal retrieval under caption perturbation.} Tab.~\ref{tab:robustness_results_refined} presents zero-shot long-caption cross-modal retrieval results under caption perturbations. The table reports the relative performance change (in \%). HyFL-CLIP (Ours) consistently outperforms other baselines across all datasets and exhibits the smallest performance drop. This indicates that the hierarchical entailment loss enables the model to better capture part–whole semantic relationships and remain robust to perturbations that preserve the underlying semantic meaning. Full results for all datasets are in the supplementary material.

\vspace{-2cm}
\begin{table}[H]
\centering
\caption{\textbf{Comparison of zero-shot long-caption cross-modal retrieval under caption perturbation.} Under caption perturbations, other models exhibit large performance drops, while our method maintains strong performance across all types and degrees of perturbations. The small numbers shown beside each score indicate the percentage performance differences (\%) relative to the original score. Notation follows Tab.~\ref{tab:longcaptionretrieval} ($*$: our implementation; $\dagger$: author-provided checkpoints).
}
\label{tab:robustness_results_refined}

\small 
\addtolength{\tabcolsep}{-1.5pt}

\begin{tabularx}{\columnwidth}{l *{5}{>{\centering\arraybackslash}X}}
\toprule
\multirow{2}{*}{Model} 
& Word & Sent. & Order & \multicolumn{2}{c}{Random} \\
& Dropout & Removal & Shuffling & \multicolumn{2}{c}{Subsampling} \\

\cmidrule(lr){2-2} 
\cmidrule(lr){3-3} 
\cmidrule(lr){4-4} 
\cmidrule(lr){5-6}

& $p=0.5$ & first & random & $n=2$ & $n=3$ \\ 
\midrule

Long-CLIP$^\dagger$~\cite{zhang2024long} 
& \mbox{48.88 \scriptsize\textcolor{blue!75}{$\downarrow$35.81}}
& \mbox{55.41 \scriptsize\textcolor{blue!75}{$\downarrow$19.45}}
& \mbox{61.79 \scriptsize\textcolor{blue!75}{$\downarrow$3.46}}
& \mbox{26.50 \scriptsize\textcolor{blue!75}{$\downarrow$91.89}}
& \mbox{31.29 \scriptsize\textcolor{blue!75}{$\downarrow$79.89}} \\

HiMo-CLIP*~\cite{wu2025himo} 
& \mbox{\underline{58.75} \scriptsize\textcolor{blue!75}{$\downarrow$27.82}}
& \mbox{\underline{64.44} \scriptsize\textcolor{blue!75}{$\downarrow$17.42}}
& \mbox{72.94 \scriptsize\textcolor{blue!75}{$\downarrow$1.88}}
& \mbox{\underline{28.24} \scriptsize\textcolor{blue!75}{$\downarrow$83.58}}
& \mbox{\underline{34.84} \scriptsize\textcolor{blue!75}{$\downarrow$71.52}} \\

FineLIP*~\cite{asokan2025finelip} 
& \mbox{58.70 \scriptsize\textcolor{blue!75}{$\downarrow$26.59}}
& \mbox{64.18 \scriptsize\textcolor{blue!75}{$\downarrow$16.25}}
& \mbox{\underline{73.41} \scriptsize\textcolor{ForestGreen!75}{$\uparrow$1.17}}
& \mbox{27.20 \scriptsize\textcolor{blue!75}{$\downarrow$86.05}}
& \mbox{33.41 \scriptsize\textcolor{blue!75}{$\downarrow$74.32}} \\

\rowcolor{gray!15}
\textbf{HyFL-CLIP (Ours)}
& \mbox{\textbf{70.20} \scriptsize\textcolor{blue!75}{$\downarrow$9.21}}
& \mbox{\textbf{68.22} \scriptsize\textcolor{blue!75}{$\downarrow$12.69}}
& \mbox{\textbf{76.16} \scriptsize\textcolor{ForestGreen!75}{$\uparrow$1.27}}
& \mbox{\textbf{32.55} \scriptsize\textcolor{blue!75}{$\downarrow$75.36}}
& \mbox{\textbf{39.05} \scriptsize\textcolor{blue!75}{$\downarrow$63.94}} \\

\bottomrule
\end{tabularx}
\end{table}

\noindent
\begin{minipage}[t]{0.45\linewidth}
\vspace{0pt}
\textbf{Zero-shot short caption cross-modal retrieval.}
Tab.~\ref{tab:shortcaptionretrieval} presents the results of zero-shot short-caption cross-modal retrieval on COCO and Flickr30k~\cite{lin2014microsoft,plummer2015flickr30k} datasets. These results indicate that HyFL-CLIP successfully distills the pre-trained text-image similarity geometry while being optimized in hyperbolic space, without sacrificing its base long-caption retrieval performance.
\end{minipage}
\hfill
\begin{minipage}[t]{0.50\linewidth}
\vspace{0pt}
\centering
\captionsetup{type=table,aboveskip=0pt,belowskip=2pt}
\caption{\textbf{Comparison of zero-shot short-caption cross-modal retrieval.} 
HyFL-CLIP achieves comparable or better short-caption retrieval performance than Euclidean baselines. 
Notation follows Tab.~\ref{tab:longcaptionretrieval} ($*$: our implementation; $\dagger$: author-provided checkpoints).}
\label{tab:shortcaptionretrieval}

\setlength{\tabcolsep}{4pt}
\resizebox{\linewidth}{!}{ 
\begin{tabular}{l|lcccc}
\toprule
\multicolumn{2}{c}{} & \multicolumn{2}{c}{COCO~\cite{lin2014microsoft}} & \multicolumn{2}{c}{Flickr30k~\cite{plummer2015flickr30k}} \\ 
\cmidrule(lr){3-4} \cmidrule(lr){5-6}
\multicolumn{2}{c}{} & I2T & T2I & I2T & T2I \\ 
\midrule
\multirow{5}{*}{\rotatebox{90}{ViT-B/16}} 
& Long-CLIP$^\dagger$~\cite{zhang2024long} & 57.26 & 40.38 & 47.19 & 33.16 \\
& TULIP~\cite{najdenkoska2024tulip} & 56.80 & 40.70 & 46.10 & \textbf{35.20} \\
& HiMo-CLIP*~\cite{wu2025himo} & \textbf{60.84} & \underline{40.71} & 50.27 & \underline{34.02} \\
& FineLIP*~\cite{asokan2025finelip} & \underline{58.32} & 40.05 & \textbf{52.23} & 33.85 \\
& \cellcolor{gray!15}\textbf{HyFL-CLIP (Ours)} 
& \cellcolor{gray!15}{58.20} 
& \cellcolor{gray!15}\textbf{41.50} 
& \cellcolor{gray!15}\underline{50.60} 
& \cellcolor{gray!15}\textbf{35.20} \\ 
\bottomrule
\end{tabular}
}
\end{minipage}

\begin{table}[H]
\centering
\caption{\textbf{Comparison of text-to-text intra modality retrieval.} Intra-modal retrieval performance across five datasets (Values in \%). Bold indicates the best, and underlined indicates the second-best. Notation follows Tab.~\ref{tab:longcaptionretrieval} ($*$: our implementation; $\dagger$: author-provided checkpoints).}
\label{tab:intramodality_final}
\resizebox{\columnwidth}{!}{ 
\begin{tabular}{lcccccccccc}
\toprule
 & \multicolumn{2}{c}{Flickr30K~\cite{plummer2015flickr30k}} & \multicolumn{2}{c}{COCO~\cite{lin2014microsoft}} & \multicolumn{2}{c}{nocaps~\cite{agrawal2019nocaps}} & \multicolumn{2}{c}{IMDB~\cite{maas2011learning}} & \multicolumn{2}{c}{20News~\cite{lang1995newsweeder}} \\ \cmidrule(lr){2-3} \cmidrule(lr){4-5} \cmidrule(lr){6-7} \cmidrule(lr){8-9} \cmidrule(lr){10-11}
Model & mAP & Pr@R & mAP & Pr@R & mAP & Pr@R & mAP & Pr@R & mAP & Pr@R \\ \midrule
Long-CLIP$^\dagger$~\cite{zhang2024long} & 52.31 & 47.56 & 27.51 & 24.58 & 37.16 & 35.47 & 51.88 & \underline{50.41} & 26.78 & 30.40 \\
HiMo-CLIP*~\cite{wu2025himo} & \underline{58.02} & \underline{52.66} & \underline{29.36} & \underline{26.03} & \underline{40.24} & \underline{38.08} & \textbf{52.65} & \textbf{50.59} & \underline{35.49} & \underline{37.61} \\
FineLIP*~\cite{asokan2025finelip} & 52.55 & 47.78 & 24.87 & 22.26 & 34.10 & 33.15 & 51.78 & 50.40 & 17.65 & 21.24 \\
\rowcolor{gray!15}
\textbf{HyFL-CLIP(Ours)} & \textbf{60.63} & \textbf{54.98} & \textbf{30.69} & \textbf{27.12} & \textbf{41.16} & \textbf{38.56} & \underline{52.60} & \textbf{50.59} & \textbf{36.69} & \textbf{38.32} \\ \bottomrule
\end{tabular}
} 
\end{table}

\begin{figure}[H]
  \centering
  \includegraphics[width=\linewidth]{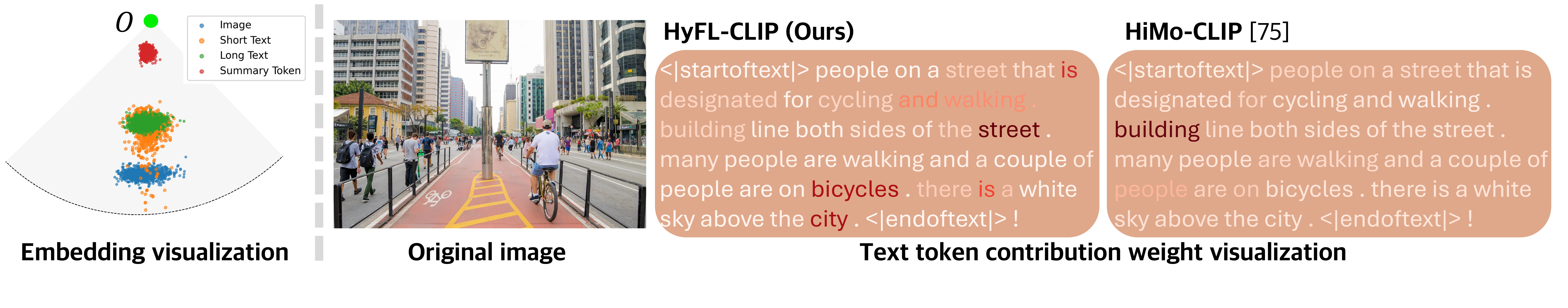}
  \caption{\textbf{Embeddings and token weight visualization.} 
  We visualize the embedding distribution of text summary token (Einstein midpoint), text, and image using HoroPCA~\cite{chami2021horopca}. We also compare text token contribution weights based on their similarity to the image.}
  \label{fig:embeddinganalysis}
\end{figure}

\begin{figure}[H]
  \centering
  \includegraphics[width=\linewidth]{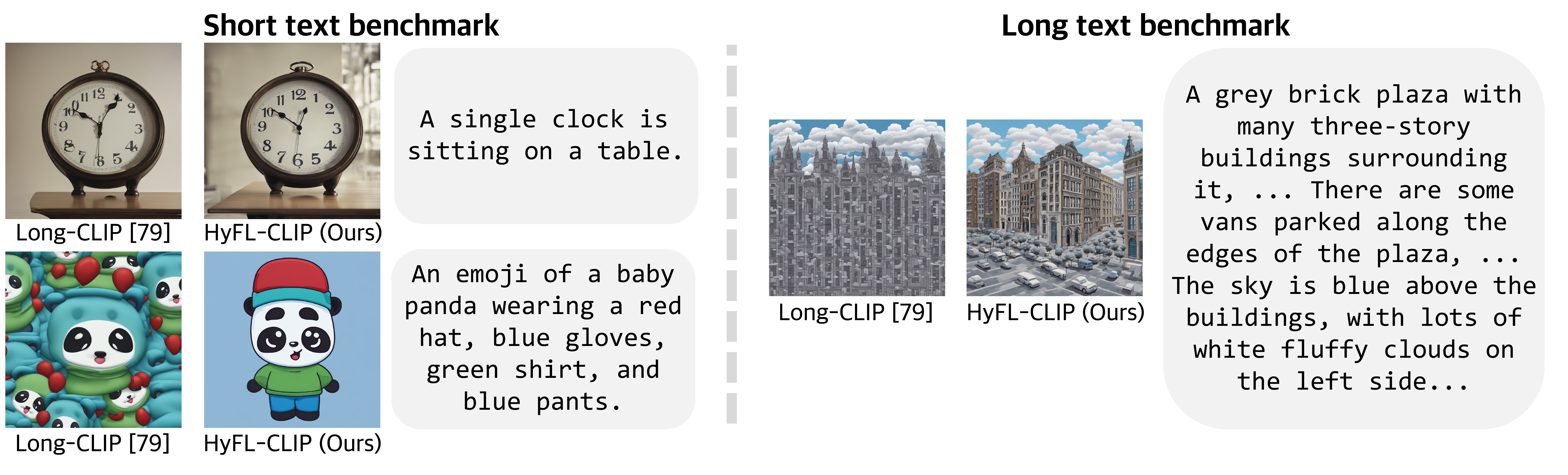}
  \caption{\textbf{SDXL text-to-image generation results.} We replace the original text encoder with ours. Hyperbolic embeddings are mapped to the Euclidean space using the logarithmic map transport. Our model captures finer details compared to the baseline.}
  \label{fig:sdxl_viz}
\end{figure}

\subsubsection{Text-to-text intra modality retrieval.} 
Tab.~\ref{tab:intramodality_final} presents the results of text-to-text intra-modality retrieval across five datasets. HyFL-CLIP (Ours) consistently achieves superior performance, demonstrating its strong capability in capturing semantic relationships within textual representations. By leveraging hierarchical entailment within the same modality, HyFL-CLIP better captures semantic abstraction between local textual components and global representations, leading to improved retrieval performance across diverse datasets.

\subsubsection{Text-to-image generation.}
Fig.~\ref{fig:sdxl_viz} presents qualitative text-to-image generation results using SDXL~\cite{podell2023sdxl} conditioned on both short captions from DrawBench~\cite{saharia2022photorealistic} and long captions from Long-DCI~\cite{najdenkoska2024tulip}. As shown in Fig.~\ref{fig:sdxl_viz}, our model captures finer-grained details, such as the number of clocks and subtle attributes of the panda, compared to the baseline model. When long prompts are used, Long-CLIP~\cite{zhang2024long} often ignores or corrupts parts of the description, whereas our model better preserves the full semantic context and generates more faithful results. These results indicate that our framework successfully distills the Euclidean similarity structure to produce more expressive hyperbolic embeddings.

\subsection{Ablation studies}
We conduct an ablation study in ViT-L/14 to evaluate the contribution of each component in our framework. First, removing $\mathcal{L}_{\text{ent}}$ reduces retrieval accuracy from 69.8\% to 69.3\%, performance averaged across all benchmarks~\cite{onoe2024docci, urbanek2024picture, najdenkoska2024tulip, zhang2024long, lin2014microsoft, plummer2015flickr30k}. Next, we further remove $\mathcal{L}_{\text{distill}}$, which results in 68.3\% performance. Full results are in the supplementary material. Fig.~\ref{fig:embeddinganalysis} shows embedding visualizations of our final model and a comparison of token contribution weights. HyFL-CLIP (Ours) assigns higher weights to semantically meaningful tokens such as street, bicycles, and city, which directly correspond to the key visual elements in the image.

\section{Conclusion}
We propose HyFL-CLIP, a framework that enables robust long-context understanding in CLIP by distilling the well-established text–image similarity geometry of a pre-trained Euclidean CLIP model into hyperbolic space. By leveraging hyperbolic geometry, HyFL-CLIP addresses the limitations of Euclidean contrastive learning by modeling hierarchical and part–whole relationships between global captions and their constituent elements. This design encourages the model to capture semantic relationships between whole captions and their constituent parts, leading to more robust representations under caption perturbations. Extensive experiments on long- and short-caption cross-modal retrieval as well as text-to-text intra-modality retrieval demonstrate state-of-the-art performance, highlighting the effectiveness of hyperbolic hierarchical representations for long-context understanding.

\section*{Acknowledgements}
This work was supported in part by Institute of Information \& communications Technology Planning \& Evaluation (IITP) grants funded by the Korea government(MSIT) [No.RS-2021-II211343, Artificial Intelligence Graduate School Program (Seoul National University) / No.RS-2025-02314125, Effective Human-Machine Teaming With Multimodal Hazy Oracle Models], the National Research Foundation of Korea(NRF) grants funded by the Korea government(MSIT) (Nos. RS-2022-NR067592, RS-2025-02263628), the AI Computing Infrastructure Enhancement (GPU Rental Support) User Support Program funded by the Ministry of Science and ICT (MSIT), Republic of Korea (No. RQT-25-120066), the BK21 FOUR program of the Education and Research Program for Future ICT Pioneers, Seoul National University, AI-Bio Research Grant through Seoul National University and the AI Seoul Tech Research Support Program of the Seoul Future Foundation.


\clearpage
\appendix

\renewcommand{\thefigure}{S\arabic{figure}}
\renewcommand{\thetable}{S\arabic{table}}
\renewcommand{\thesection}{S\arabic{section}}
\setcounter{figure}{0}
\setcounter{table}{0}
\setcounter{section}{0}


\title{Supplementary Material for \\HyFL-CLIP: Hyperbolic Fine-Tuning of CLIP for Robust Long-Context Understanding}
\titlerunning{Hyperbolic Fine-Tuning of CLIP for Robust Long-Context Understanding}
\author{
Ji Ha Jang$^{1,*}$\orcidlink{0009-0003-6310-8262},\qquad
Hayeon Kim$^{1,*}$\orcidlink{0009-0008-7461-9750},\qquad
Chulwon Lee$^{2}$\orcidlink{0009-0003-4668-3644},\qquad \\
Junghun James Kim$^{2}$\orcidlink{0009-0001-8327-9662},\qquad  
Se Young Chun$^{1,2,3,\dagger}$\orcidlink{0000-0001-8739-8960}}

\authorrunning{JH Jang, H Kim \textit{et al.}}

\institute{$^1$Dept. of Electrical and Computer Engineering, $^2$IPAI, $^3$INMC \& AIIS, \\
Seoul National University, Republic of Korea \\ 
\email{\{jeeit17, khy5630, chul0e, jonghean12, sychun\}@snu.ac.kr}}
\maketitle
\def\thefootnote{*}\footnotetext{Authors contributed equally. $^{\dagger}$ Corresponding author.}


\section{Additional Experimental Details\label{exep_detail}}

\subsection{Zero-shot long/short caption cross modality retrieval}
\subsubsection{Datasets.} We use four datasets for zero-shot long-caption retrieval following prior works~\cite{zhang2024long, wu2025himo, najdenkoska2024tulip, asokan2025finelip, wang2025fix, feng2025retaining}. Descriptions of Connected and Contrasting Images (DOCCI)~\cite{onoe2024docci} contains approximately $15$k images paired with long, human-written English descriptions, with captions averaging around $141.5$ tokens. The annotations emphasize fine-grained visual details, including spatial relationships between objects, counting, and text appearing in the scene. Urban-1k, introduced in~\cite{zhang2024long}, extends the earlier Urban-200 dataset proposed in the same work. The dataset contains $1$k images with long captions averaging $131.4$ tokens. It is constructed by selecting visually similar images from the Visual Genome dataset~\cite{krishna2017visual}, after which long descriptive captions are generated using GPT-4V~\cite{achiam2023gpt} to provide detailed scene descriptions. Densely Captioned Images (DCI)~\cite{urbanek2024picture} consists of $7{,}805$ natural images annotated with dense, human-written descriptions aligned with segmentation masks. The captions are highly detailed, averaging approximately $174.2$ tokens per image, and are designed to support fine-grained vision–language understanding. Long-DCI~\cite{najdenkoska2024tulip} is derived from DCI and contains about $7$k images paired with long human-annotated captions, with an average length of $200$ tokens.

For short-caption retrieval, we follow prior work~\cite{zhang2024long, wu2025himo, najdenkoska2024tulip, asokan2025finelip, wang2025fix, feng2025retaining} and evaluate on the COCO2017 5k validation split~\cite{lin2014microsoft} and the full Flickr30k dataset~\cite{plummer2015flickr30k}, where the average caption lengths are approximately $13.5$ and $15.8$ tokens, respectively.

\subsubsection{Evaluation protocol.} 
We evaluate cross-modal retrieval using top-1 accuracy. All methods are evaluated using the ViT-B/16 backbone for fair comparison. Given a dataset of paired images and captions, we first encode all captions and images using the text and image encoders of the model to obtain their corresponding embeddings. Text inputs longer than $248$ tokens are truncated during tokenization. The resulting features are then mapped to the hyperbolic space, where similarity between image and text embeddings is computed using the Lorentzian inner product with the learned curvature parameter. Unlike CLIP~\cite{Radford2021LearningTV}, we do not apply feature normalization, and similarities are computed directly in the hyperbolic space. For both image-to-text and text-to-image retrieval, each query is compared with all candidates from the other modality, and the item with the highest similarity score is retrieved. A prediction is counted as correct if the retrieved item matches the ground-truth pair.

For CLIP-based baselines~\cite{zhang2024long, wu2025himo, najdenkoska2024tulip, asokan2025finelip, wang2025fix, feng2025retaining}, image and text embeddings are $\ell_2$-normalized and similarities are computed using cosine similarity instead of the Lorentzian inner product. 
\label{subsec:zero-shot long-caption cross-modal retrieval}

\subsection{Zero-shot long-caption cross-modal retrieval under caption perturbation}\label{sec:perturb_evaluation_protocol}
\subsubsection{Caption perturbation.} To evaluate robustness to caption perturbations, we generate modified captions using four types of perturbations. Word dropout randomly removes a fraction of words from the caption. In our experiments, we use a dropout probability of $p=0.5$, meaning that approximately half of the words are randomly removed. Sentence removal deletes the first sentence of the caption, which often contains a high-level summary of the scene, to simulate the absence of this summary-level information. Sentence order shuffling randomly permutes the order of sentences within the caption while keeping the sentence contents unchanged. Random subsampling constructs a shortened caption by randomly selecting a subset of sentences from the original caption. We use $n=2$ and $n=3$ sentences in our experiments. 

\subsubsection{Evaluation protocol.} The retrieval evaluation follows the protocol described in Sec~\ref{subsec:zero-shot long-caption cross-modal retrieval}. For each perturbation setting, the experiments are repeated five times for all datasets to account for randomness in the perturbation process, and the reported results are averaged across runs. The drop rates reported in the main paper indicate the relative performance decrease with respect to each model’s original retrieval performance without perturbations. 

\subsection{Text-to-text intra modality retrieval}
\subsubsection{Evaluation protocol.} To evaluate whether the learned text embeddings capture semantic consistency between full captions and their constituent parts, we additionally perform text-to-text intra-modality retrieval following~\cite{mistretta2025cross}. In this setting, captions are divided into a query set and a gallery set, where the gallery serves as the retrieval database. All captions are encoded using the text encoder to obtain embeddings. For our model, similarities between embeddings are computed using the Lorentzian inner product, while Euclidean baselines~\cite{zhang2024long,wu2025himo,asokan2025finelip} compute similarities using cosine similarity. For each query caption, similarities are computed against all captions in the gallery and the captions are ranked according to their similarity scores. 

Retrieval performance is evaluated using mean Average Precision (mAP) and Precision@R (Pr@R). The mAP measures the average precision over the ranked retrieval results, reflecting the overall ranking quality. Precision@R computes the precision at rank R, where R denotes the number of relevant captions for each query and therefore varies across queries. Both metrics are computed per query and averaged across all queries. All methods are evaluated using the ViT-B/16 backbone.

\subsection{Text-to-image generation} 
Our Stable Diffusion XL (SDXL)~\cite{podell2023sdxl} implementation preserves the original two-stream text-conditioning interface, where token-level embeddings are formed by concatenating an OpenCLIP-L/14 branch and an OpenCLIP-bigG branch, with pooled embeddings retained from the OpenCLIP-bigG branch. Token-level text features from HyFL-CLIP (Ours) and LongCLIP~\cite{zhang2024long} are injected only into the compatible channels of the OpenCLIP-L/14 branch via $\alpha$-interpolation, while the remaining channels and the second branch are kept unchanged. This branch-local design isolates the effect of the injected long-context text representations while keeping the rest of the generation pipeline fixed, following prior controlled studies on text representations in text-to-image generation~\cite{li2024textcraftor}. For fair comparison, all methods are evaluated under the same protocol with a ViT-B/16 backbone. Implementation results with a ViT-L/14 backbone are also reported in Sec.~\ref{sec:SDXL suppl}.

\section{Additional Pipeline Details\label{exep_detail}}  

\subsection{Model architecture} We initialize our model from the pretrained weights of OpenCLIP~\cite{ilharco_gabriel_2021_5143773}. The text encoder follows the CLIP~\cite{Radford2021LearningTV} architecture and consists of 12-layer Transformer~\cite{vaswani2017attention} with a hidden dimension of $512$. The maximum input length is set to $248$ tokens. Following prior works~\cite{zhang2024long, wu2025himo, najdenkoska2024tulip, asokan2025finelip, wang2025fix, feng2025retaining}, we preserve the positional embeddings of the first $20$ tokens and apply interpolation only to the remaining positions, allowing longer input sequences while maintaining the well-trained positional structure of CLIP.  
\begin{equation}
PE^*(pos)=
\begin{cases}
PE(pos), & pos \le 20 \\
(1-\alpha)\, PE\!\left(\left\lfloor \frac{pos}{r} \right\rfloor\right)
+
\alpha\, PE\!\left(\left\lceil \frac{pos}{r} \right\rceil\right),
& \text{otherwise}
\end{cases}
\end{equation}

where $PE(pos)$ denotes the original positional embedding at position $pos$, and

\begin{equation}
\alpha = \frac{pos \bmod r}{r}.
\end{equation}
Here, $\alpha \in [0,1]$ controls the interpolation weight between the two neighboring positions, and $r = 4$ in our case. This results in an extended positional embedding length of $248$ tokens from the original $77$ tokens.

For images, we adopt a Vision Transformer~\cite{dosovitskiy2020image} and experiment with two capacity configurations, ViT-B/16 and ViT-L/14, using a patch size of 16 and 14, respectively. 

\subsection{Model initialization and hyperparameter setting} We parameterize the curvature in the Lorentz space and treat it as a learnable parameter. The curvature is initialized with $\kappa = 1.0$ and converges to $0.9994$ after training. Following prior works~\cite{desai2023hyperbolic, pal2024compositional, ramasinghe2024accept}, we apply learnable scaling factors to image and text vectors, setting $c_{\text{img}} = c_{\text{txt}} = \frac{1}{\sqrt{512}}$ for numerical stability.

For short-text guided cross-manifold similarity distillation, we set $\tau_E = \tau_H = 0.005$ in Eq.~(12). In the hyperbolic geodesic contrastive loss (Eq.~(14) and Eq.~(15)), the temperature parameter $\tau_c$ is set to $0.07$. For hierarchical entailment via Einstein midpoint aggregation, we follow the standard hyperbolic entailment cone formulation and set $K = 1$ in Eq.~(19). The scaling parameter $\eta$ is set to $1.2$ in Eq.~(20). Finally, we set $\lambda_1 = 0.1$ in Eq.~(16), and $\lambda_2 = 0.05$, $\lambda_3 = 0.1$, $\lambda_4 = 0.1$ in Eq.~(22). A sensitivity analysis of the hyperparameters is provided in Sec.~\ref{sec:hyperparameter_sensitivity}.

\subsection{Training details} 

\subsubsection{Optimizer.} Our model is trained for $2$ epochs using four A100 GPUs with a global batch size of $1024$. We employ the AdamW optimizer~\cite{loshchilov2017decoupled}, setting $\beta_1=0.9, \beta_2=0.999$, and a weight decay of $2.5 \times 10^{-2}$. We adopt a cosine learning-rate scheduler~\cite{loshchilov2016sgdr} with a learning rate of $10^{-5}$, with a 200-step linear warm-up period.

\subsubsection{Computational overhead.} We compare the computational overhead of HyFL-CLIP (Ours) with several baselines~\cite{najdenkoska2024tulip, asokan2025finelip, wu2025himo, zhang2024long}. All comparisons are conducted using the ViT-B/16 backbone. Tab.~\ref{tab:flops} reports the total computational cost in terms of FLOPs across the full training schedule. Overall, our model requires less computational overhead than the compared methods.

Tulip~\cite{najdenkoska2024tulip} employs a distillation-based training strategy. Specifically, it performs distillation from a teacher OpenCLIP model for 20 epochs followed by 1 epoch of fine-tuning, which we denote as `20+1' epochs in the table. Following the convention used in prior work, the computational cost of the teacher model during distillation is not included in the reported FLOPs.

\begin{table}[t]
\centering
\setlength{\tabcolsep}{10pt}
\renewcommand{\arraystretch}{1.25}
\caption{\textbf{Comparison of total computational cost (FLOPs) across different methods.} We report the total computational cost of HyFL-CLIP (Ours) and several baseline methods. Overall, our model requires less computational overhead than the compared methods.} 
\label{tab:flops}
\begin{tabular}{lcc}
\hline
Model & Epochs & Total compute (FLOPs) \\
\hline
Long-CLIP~\cite{zhang2024long} & 1 & $2.66 \times 10^{17}$ \\
HiMo-CLIP~\cite{wu2025himo} & 10 & $1.96 \times 10^{18}$ \\
FineLIP~\cite{asokan2025finelip} & 6 & $1.19 \times 10^{18}$ \\
Tulip~\cite{najdenkoska2024tulip} & 20 (+1) & $1.58 \times 10^{18}$ \\
\rowcolor{gray!15}
HyFL-CLIP (Ours) & 2 & $5.34 \times 10^{17}$ \\
\hline
\end{tabular}
\end{table}

\begin{figure}[t]
\centering
  \includegraphics[width=0.75\linewidth]{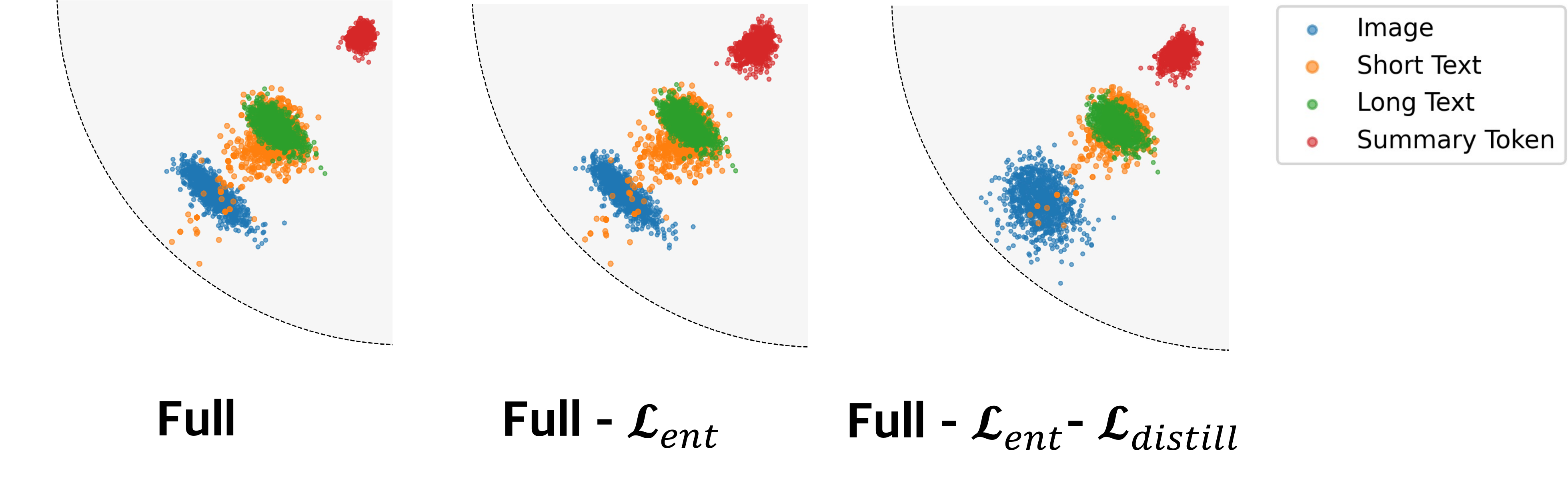}
  \caption{\textbf{Qualitative results of the full ablation study.} We visualize the embedding distributions of images (Image), short texts (Short Text), corresponding long texts (Long Text), and Einstein midpoints of long texts (Summary Token) using HoroPCA~\cite{chami2021horopca} for the full model and ablated variants without $\mathcal{L}_{ent}$ and $\mathcal{L}_{distill}$ in ShareGPT4V dataset~\cite{chen2024sharegpt4v}.
  }
  \label{fig:abl_full}
\end{figure}

\begin{table}[H]
\centering
\caption{\textbf{Quantitative results of the full ablation study.}
We analyze the contribution of each objective by removing $\mathcal{L}_{ent}$ and $\mathcal{L}_{distill}$ from the full model. 
All ablations are conducted using the ViT-L/14 backbone. 
The experiments are performed across multiple datasets.}

\label{tab:ablation_objectives}
\setlength{\tabcolsep}{6pt}
\renewcommand{\arraystretch}{1.1}

\resizebox{\columnwidth}{!}{
\begin{tabular}{lcccccc}
\toprule

& \multicolumn{2}{c}{Full - $\mathcal{L}_{\text{ent}}$ - $\mathcal{L}_{\text{distill}}$} 
& \multicolumn{2}{c}{Full - $\mathcal{L}_{\text{ent}}$} 
& \multicolumn{2}{c}{Full} \\

\cmidrule(lr){2-3}
\cmidrule(lr){4-5}
\cmidrule(lr){6-7}

Dataset & I2T & T2I & I2T & T2I & I2T & T2I \\

\midrule
Urban-1k~\cite{zhang2024long}      & 93.80 & 93.70 & 95.00 & 94.80 & 94.60 & 94.30 \\
DOCCI~\cite{onoe2024docci}         & 82.35 & 82.96 & 82.18 & 85.41 & 82.12 & 85.39 \\
DCI~\cite{urbanek2024picture}      & 74.64 & 73.99 & 73.69 & 75.89 & 74.74 & 76.19 \\
Long-DCI~\cite{najdenkoska2024tulip} & 62.00 & 61.42 & 61.64 & 63.55 & 61.92 & 63.93 \\

\midrule
COCO~\cite{lin2014microsoft}       & 42.77 & 61.78 & 41.56 & 61.06 & 45.56 & 61.56 \\
Flickr30k~\cite{plummer2015flickr30k} & 35.78 & 54.92 & 40.80 & 55.70 & {41.18} & {56.32} \\

\bottomrule
\end{tabular}
}

\end{table}
\section{Additional Ablation Results}
\subsection{Full ablation results}
Tab.~\ref{tab:ablation_objectives} and Fig.~\ref{fig:abl_full} present the full ablation results, highlighting the complementary effects of the two objectives, $\mathcal{L}_{ent}$ and $\mathcal{L}_{distill}$.

Removing $\mathcal{L}_{ent}$ slightly degrades retrieval performance on long-context datasets. 
As shown in Fig.~\ref{fig:abl_full}, the alignment between long-text embeddings and their corresponding summary tokens becomes weaker, resulting in a more dispersed distribution of long-text representations. This suggests that $\mathcal{L}_{ent}$ may help align long texts with their summary tokens and improve the stability of the long-context representation space.

In contrast, removing $\mathcal{L}_{distill}$ results in a noticeable drop in retrieval performance, particularly on datasets with short captions. The corresponding visualization shows weakened alignment between image and text embeddings, where the image cluster drifts away from the text clusters. This suggests that $\mathcal{L}_{distill}$ helps preserve the cross-modal semantic structure inherited from the pretrained CLIP representation.

\begin{table}[H]
\centering
\caption{\textbf{Hyperparameter sensitivity analysis.}
We vary each hyperparameter ($\lambda_1$–$\lambda_4$) over a range of values while keeping the others fixed and report the resulting performance. The results show that the performance remains stable across a broad range of hyperparameter values, indicating that our method is not sensitive to precise hyperparameter tuning.}
\label{tab:lambda_sensitivity}

\footnotesize
\begin{tabular}{lcc>{\columncolor{gray!15}}c cc}
\toprule

\multirow{2}{*}{$\lambda_1$}
& 0.01
& 0.05
& 0.10
& 0.15
& 0.20 \\
\cmidrule(lr){2-6}
& 74.58/75.35 & 74.63/75.95 & 75.18/75.70 & 75.15/75.93 & 75.35/75.75 \\
\midrule

\multirow{2}{*}{$\lambda_2$}
& 0.01
& 0.03
& 0.05
& 0.07
& 0.09 \\
\cmidrule(lr){2-6}
& 75.13/75.35 & 75.08/75.83 & 75.18/75.70 & 75.10/75.95 & 75.03/76.05 \\
\midrule

\multirow{2}{*}{$\lambda_3$}
& 0.01
& 0.05
& 0.10
& 0.15
& 0.20 \\
\cmidrule(lr){2-6}
& 75.05/75.83 & 75.33/76.03 & 75.18/75.70 & 75.13/75.65 & 75.18/75.88 \\
\midrule

\multirow{2}{*}{$\lambda_4$}
& 0.01
& 0.05
& 0.10
& 0.15
& 0.20 \\
\cmidrule(lr){2-6}
& 75.10/75.88 & 75.15/75.85 & 75.18/75.70 & 75.23/75.65 & 75.03/75.85 \\

\bottomrule
\end{tabular}

\end{table}

\subsection{Hyperparameter sensitivity test}\label{sec:hyperparameter_sensitivity} We analyze the sensitivity of the hyperparameters $\lambda_1$, $\lambda_2$, $\lambda_3$, and $\lambda_4$. For each experiment, one hyperparameter is varied while the remaining ones are fixed to their default values. Specifically, when evaluating $\lambda_1$, we fix $\lambda_2=0.05$, $\lambda_3=0.1$, and $\lambda_4=0.1$. Similarly, $\lambda_2$, $\lambda_3$, and $\lambda_4$ are analyzed by varying each parameter individually while keeping the others fixed. Each entry of Tab.~\ref{tab:lambda_sensitivity} reports the corresponding I2T / T2I retrieval performance. The results show that the performance remains stable across different hyperparameter settings, indicating that our method is not sensitive to precise hyperparameter tuning.

\section{Analysis on robustness to caption perturbations}
\subsection{Embedding analysis under caption perturbations in Hyperbolic and Euclidean spaces}\label{sec:analysis_embedding} 
\subsubsection{Embedding visualization with perturbed captions.} We visualize the embedding distributions to examine the relationship between the original text–image pairs and their perturbed texts in both Euclidean and hyperbolic spaces. Specifically, we employ t-SNE~\cite{van2008visualizing} for HiMo-CLIP~\cite{wu2025himo}, which operates in Euclidean space, and HoroPCA~\cite{chami2021horopca} for HyFL-CLIP (Ours), whose embeddings lie in hyperbolic space. 

\begin{figure}[tb]
  \includegraphics[width=\linewidth]{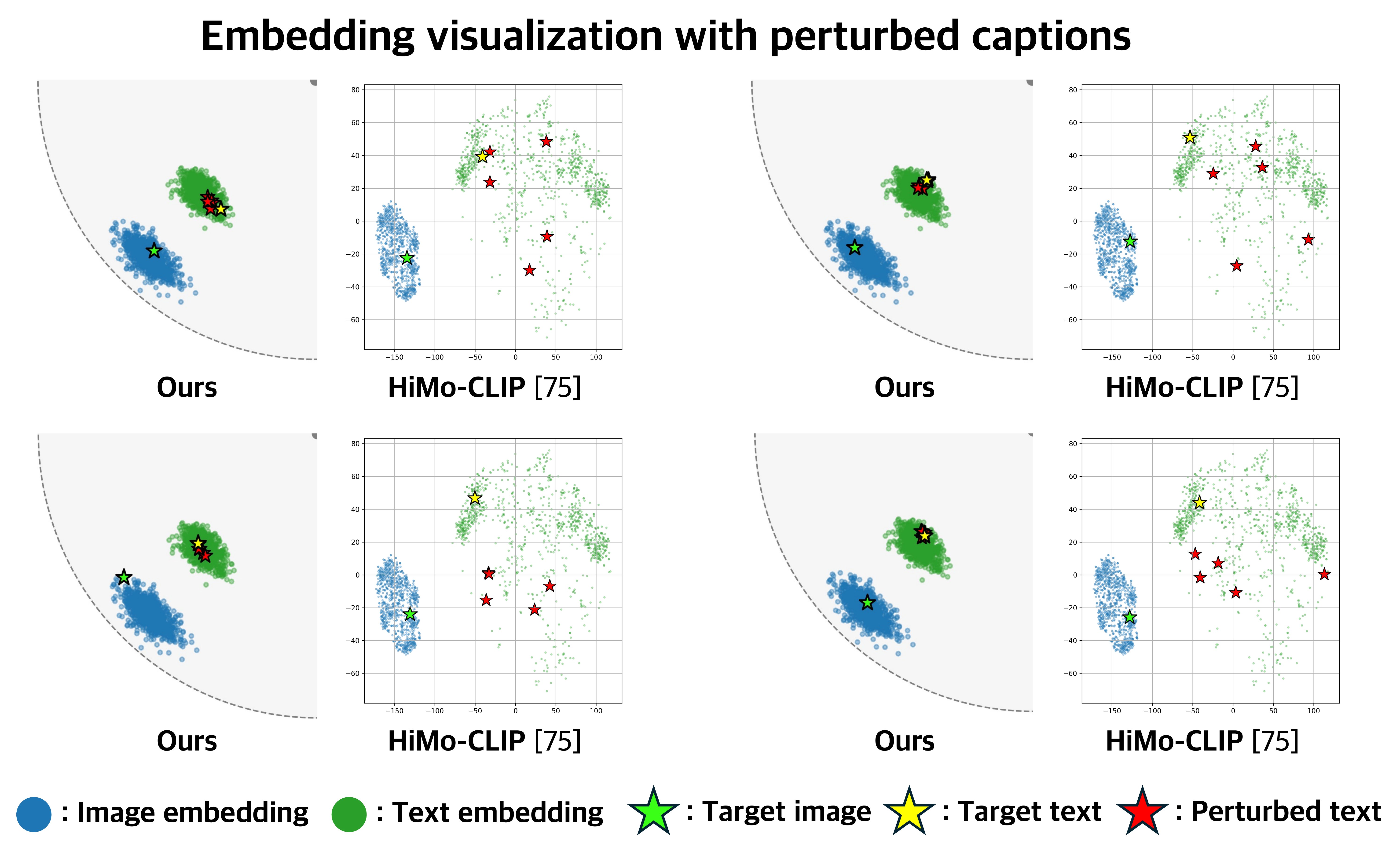}
 \caption{\textbf{Embedding visualization with perturbed captions.}
We visualize the target image and text embeddings together with captions generated by perturbing the target text embedding. Under Euclidean similarity (HiMo-CLIP~\cite{wu2025himo}), perturbed caption embeddings are widely scattered in the embedding space. In contrast, our hyperbolic representation keeps perturbed captions tightly clustered around the target text embedding, indicating more stable semantic alignment with the target image.}
  \label{fig:embedding_vis}
\end{figure}

As shown in Fig.~\ref{fig:embedding_vis}, in Euclidean space the embeddings of perturbed texts (red stars) are separated from the original target text (yellow star), indicating that even subtle variations in the caption can lead to large shifts in the embedding space under the Euclidean distance metric. In contrast, in hyperbolic space the embeddings of perturbed texts remain close to the original target text embedding. This behavior highlights the property of the hyperbolic distance metric, which better preserves semantic proximity under caption perturbations and leads to more stable alignment between images and their corresponding textual descriptions.

\subsubsection{Rank distribution of image-text similarity under caption perturbations.} We conduct an analysis to evaluate how accurately the original target image can be retrieved given various perturbed captions. Specifically, each perturbed caption is mixed with all other captions, and the captions are then ranked according to their similarity to the original target image. In Euclidean space, embeddings are typically normalized to unit length, allowing the Euclidean distance to be measured through cosine similarity. In contrast, hyperbolic space enables the use of both the entailment relationship between embeddings and the hyperbolic distance. With HiMo-CLIP~\cite{wu2025himo}, image–text similarity is ranked using Euclidean distance, as the model operates in Euclidean space. In contrast, our model ranks image–text similarity using both hyperbolic distance and the entailment relationship.

\begin{table}[t]
\centering
\setlength{\tabcolsep}{10pt}
\renewcommand{\arraystretch}{1.25}
\caption{\textbf{Comparison of ranking statistics.} We report the mean rank and rank variance of perturbed captions in the caption similarity ranking. Lower mean rank values indicate that perturbed captions are retrieved closer to the top positions, while lower rank variance indicates more stable and consistent retrieval behavior. Compared to Euclidean distance, hyperbolic distance improves both the accuracy and stability of retrieving semantically similar perturbed captions, and hyperbolic entailment exposes this relationship more clearly.}
\label{tab:rank_stats}
\begin{tabular}{lcc}
\hline
Model & Mean Rank & Rank Variance \\
\hline
HiMo-CLIP~\cite{wu2025himo} (Euclidean distance) & 184.94 & 42250.26 \\
Ours (Hyperbolic distance) & 156.99 & 31007.93 \\
Ours (Hyperbolic entailment) & 78.50 & 12666.22 \\
\hline
\end{tabular}
\end{table}

\begin{figure}[H]
  \includegraphics[width=\linewidth]{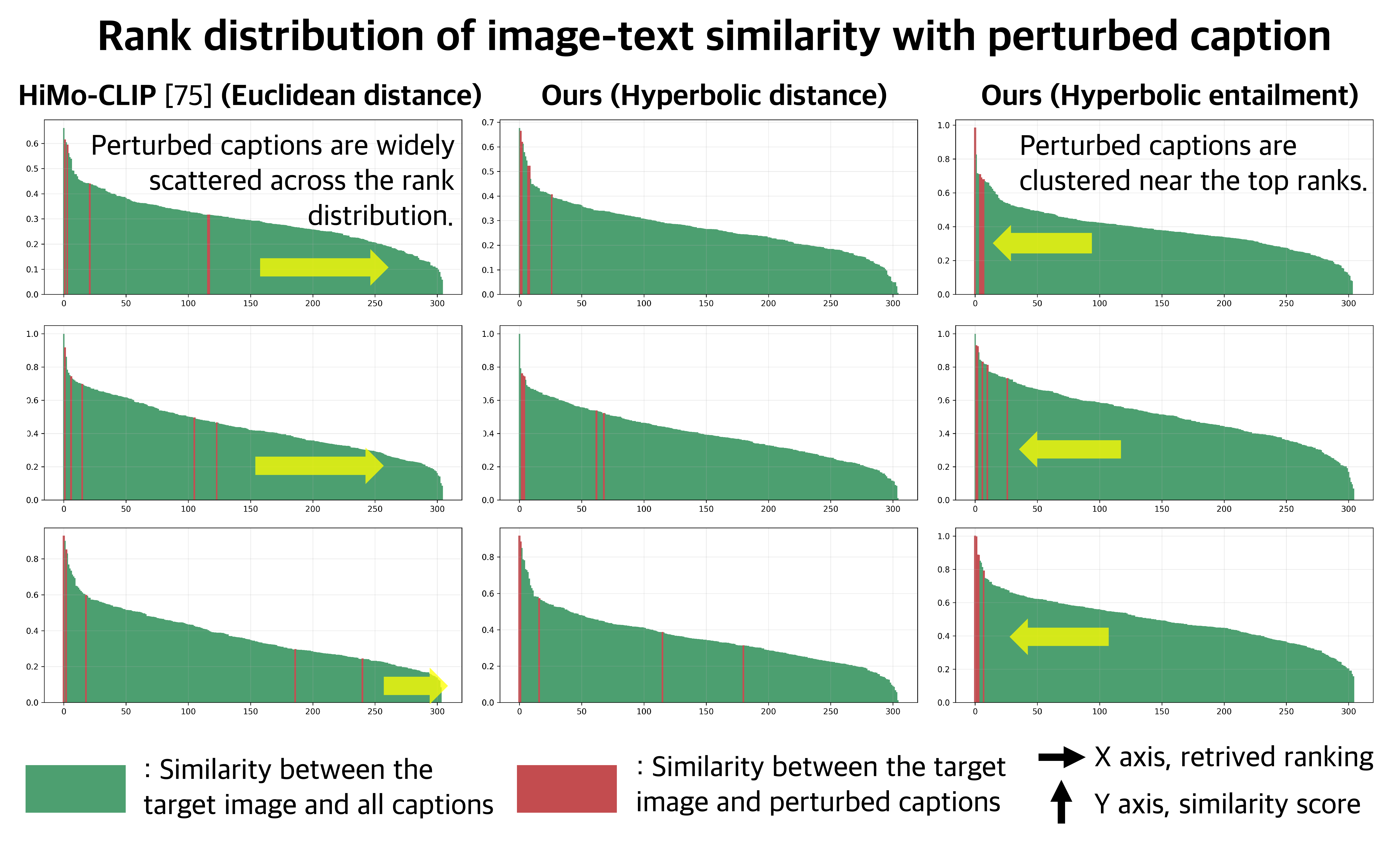}
 \caption{\textbf{Rank distribution of image–text similarity under caption perturbations.}
Captions are ranked by similarity to the target image. Green bars show similarities with all captions, while red markers indicate captions perturbed from the reference caption. 
Euclidean similarity (HiMo-CLIP~\cite{wu2025himo}) scatters perturbed captions across the ranking, whereas hyperbolic entailment clusters them near the top ranks.}
  \label{fig:ranking_distrib}
\end{figure}

Fig.~\ref{fig:ranking_distrib} and Tab.~\ref{tab:rank_stats} present the ranking results. Consistent with the observations in Fig.~\ref{fig:embedding_vis}, ranking based on Euclidean distance scatters perturbed captions widely across the rank distribution, even though they retain similar semantics to the original captions. In contrast, our hyperbolic distance produces a distribution in which perturbed captions are concentrated near the top ranks.

This trend is also reflected in the ranking statistics in Tab.~\ref{tab:rank_stats}. Here, the mean rank and rank variance are computed over the ranks of perturbed captions with respect to the target image. Compared to Euclidean distance, our hyperbolic distance yields a lower mean rank and lower rank variance, indicating that semantically similar perturbed captions are retrieved more consistently and appear closer to the top of the ranking.

This tendency becomes even more pronounced when using the hyperbolic entailment relationship. As shown in both Fig.~\ref{fig:ranking_distrib} and Tab.~\ref{tab:rank_stats}, when ranking is based on the entailment relationship between the original caption and its perturbed captions, the perturbed captions are concentrated almost entirely at the top of the ranking. Correspondingly, Tab.~\ref{tab:rank_stats} shows that the mean rank and rank variance of the perturbed captions are substantially reduced.

These two experiments demonstrate that hyperbolic space is more robust than Euclidean distance in preserving semantic relationships under perturbations, preventing semantically related representations from becoming discretely separated or drifting far from the original representation. Moreover, this tendency becomes more pronounced when using the entailment relationship, which is naturally defined in hyperbolic space.

\begin{table}[H]
\centering
\caption{\textbf{Performance degradation (\%) under caption perturbations, measured relative to each model's original (non-perturbed) performance.} This comparison evaluates the influence of each training objective by analyzing how the robustness changes when individual losses are removed.
}
\label{tab:perturbation_analysis}

\setlength{\tabcolsep}{8pt}
\renewcommand{\arraystretch}{1.2}

\begin{tabular}{lcccc}
\toprule
Dataset & w/o $\mathcal{L}_{\text{distill}}$ & w/o $\mathcal{L}_{\text{ent}}$ & Ours full \\
\midrule
Urban-1k~\cite{zhang2024long} &32.99 &36.85  &33.83  \\
DOCCI~\cite{onoe2024docci} &29.57 &33.85  &30.48  \\
DCI~\cite{urbanek2024picture} &27.26 &32.82  &29.89 \\
DCI-Long~\cite{najdenkoska2024tulip} &39.31  &39.04 &35.31  \\
\midrule
Average &32.28 &35.64  &32.38  \\
\bottomrule
\end{tabular}

\end{table}

\subsection{Impact of training objectives on caption perturbation robustness} To further investigate the observations from Sec.~\ref{sec:analysis_embedding} and identify which loss term contributes to robustness against caption perturbations, we conduct an ablation study by removing each loss term ($\mathcal{L}_{ent}$ and $\mathcal{L}_{distill}$) individually. We then measure the performance degradation under caption perturbations. The types and number of perturbations follow the same evaluation protocol used in the main experiments (Sec.~\ref{sec:perturb_evaluation_protocol}). For each model variant, the performance degradation is measured relative to its own original performance rather than the full model, and the results are averaged over all perturbation settings and datasets.

Tab.~\ref{tab:perturbation_analysis} shows that removing $\mathcal{L}_{ent}$ results in a substantially larger performance drop under caption perturbations compared to removing $\mathcal{L}_{distill}$. This indicates that $\mathcal{L}_{ent}$ plays a more important role in improving robustness to caption perturbations. This observation is consistent with the findings in Sec.~\ref{sec:analysis_embedding}, where the entailment relationship in hyperbolic space relaxes the strict one-to-one correspondence between image–text pairs and allows semantically related concepts to remain close in the embedding space, thereby improving robustness to caption perturbations.

Fig.~\ref{fig:activation_viz1}, Fig.~\ref{fig:activation_viz2}, and Fig.~\ref{fig:activation_viz3} show the visualization of text-token contribution weights when the model is trained with and without $\mathcal{L}_{ent}$. When trained with $\mathcal{L}_{ent}$, the entailment relationship encourages the Einstein midpoint of token embeddings to align with the global token representation of the long text. Consequently, tokens corresponding to different visual regions or attributes of the image are more clearly highlighted.

\begin{table}[H]
\centering
\caption{\textbf{BLIP-VQA robustness under caption perturbations.}
We measure image–text consistency using BLIP-VQA on images retrieved from perturbed captions. 
Compared with HiMo-CLIP, our method maintains higher consistency with the captions across multiple perturbation strategies.}
\label{tab:blipvqa_robustness}

\small
\addtolength{\tabcolsep}{-1.5pt}

\begin{tabularx}{\columnwidth}{l *{5}{>{\centering\arraybackslash}X}}
\toprule

\multirow{2}{*}{Model} 
& Word & \multicolumn{2}{c}{Random} & Order & Sent. \\
& Dropout & \multicolumn{2}{c}{Subsampling} & Shuffling & Removal \\

\cmidrule(lr){2-2}
\cmidrule(lr){3-4}
\cmidrule(lr){5-5}
\cmidrule(lr){6-6}

& $p=0.5$ & $n=2$ & $n=3$ & random & first \\

\midrule

GT
& \mbox{\textbf{0.4725}}
& \mbox{\textbf{0.4869}}
& \mbox{\textbf{0.4809}}
& \mbox{\textbf{0.4581}}
& \mbox{\textbf{0.4395}} \\

\rowcolor{gray!15}
HyFL-CLIP (Ours)
& \mbox{\underline{0.4112}}
& \mbox{\underline{0.4283}}
& \mbox{\underline{0.4160}}
& \mbox{\underline{0.4372}}
& \mbox{\underline{0.3997}} \\

HiMo-CLIP~\cite{wu2025himo}
& \mbox{0.3864}
& \mbox{0.3982}
& \mbox{0.3917}
& \mbox{0.4321}
& \mbox{0.3795} \\

\bottomrule
\end{tabularx}
\end{table}

\subsection{VQA-based validation of retrieval results under caption perturbations}

To evaluate whether perturbed captions remain semantically valid with respect to their paired images, we measure image–text consistency using an LLM-based VQA framework (Tab.~\ref{tab:blipvqa_robustness}). After applying caption perturbations, we use the perturbed text as input to the BLIP-VQA~\cite{huang2025t2i} to generate five questions derived from the caption. These questions are then used to evaluate the validity of three candidate images: the ground-truth (GT) image paired with the original caption, the top-1 retrieved image from our method, and the top-1 retrieved image from HiMo-CLIP. For each candidate image, we measure the True/False correctness of the answers produced by BLIP-VQA.
We conduct the evaluation on the Urban-1k dataset~\cite{zhang2024long}. For each type of caption perturbation, the experiment is repeated five times and the results are averaged. The quantitative results are summarized in Tab.~\ref{tab:blipvqa_robustness}. The results show that the original GT image maintains the highest alignment with the perturbed captions. Importantly, the images retrieved by our method also remain highly consistent with the perturbed captions, demonstrating that our model retrieves images that are still semantically valid even under caption perturbations.

\subsection{Robustness under LLM-generated hard negative perturbations} 

\subsubsection{Experimental setup.} Beyond the perturbations considered in the main paper, we further conduct robustness experiments with hard-negative perturbations to examine whether the model can correctly distinguish positive captions from negative ones even under extremely subtle changes in long-context captions. We generate hard-negative captions for the Urban-1k dataset~\cite{zhang2024long} by replacing a single word in the original captions using LLaMA-30B~\cite{touvron2023llama}, following FG-OVD benchmark~\cite{bianchi2024devil}. Content words (nouns, verbs, adjectives, and adverbs) are identified using part-of-speech tagging, and up to five of them are randomly selected per caption. Each selected word is replaced by the language model with a grammatically valid word of different meaning, generating up to five hard-negative captions per sample. An example of an LLM-generated hard-negative sample is shown in Fig.~\ref{fig:hard_neg}.

\subsubsection{Evaluation protocol.} For each model, we compute the similarity between the image and the positive caption, as well as the similarity between the image and the corresponding hard-negative caption. The accuracy is defined as the percentage of cases where the model assigns a higher similarity score to the positive caption than to the hard-negative caption. This evaluation protocol is consistent with existing benchmarks~\cite{bianchi2024devil, zhao2022vl,hsieh2023sugarcrepe}.

\begin{table}[H]
\centering
\caption{\textbf{Robustness to LLM-generated hard-negative perturbations.} Hard-negative captions are generated for the Urban-1k dataset~\cite{zhang2024long} by replacing a single word in the original captions using LLaMA-30B~\cite{touvron2023llama}. The metric measures the percentage of cases where the similarity between the image and the positive caption exceeds that between the image and the corresponding hard-negative caption. Our model achieves the highest score, indicating a stronger ability to distinguish subtle semantic differences in captions.}
\label{tab:hard_negative}

\setlength{\tabcolsep}{8pt}
\renewcommand{\arraystretch}{1.2}

\begin{tabular}{lc}
\toprule
Model & Accuracy (\%) \\
\midrule
Long-CLIP~\cite{zhang2024long} & 54.01 \\
HiMo-CLIP~\cite{wu2025himo} & 52.94 \\
FineLIP~\cite{asokan2025finelip} & 53.54 \\
\rowcolor{gray!15}
\textbf{HyFL-CLIP (Ours)} & \textbf{62.07} \\
\bottomrule
\end{tabular}

\end{table}

\subsubsection{Experimental results.}
Experimental results in Tab.~\ref{tab:hard_negative} show that HyFL-CLIP (Ours) achieves strong performance in distinguishing hard-negative captions from positive ones. While other models perform close to near-random under this challenging setting, our model maintains a clear margin. This improvement can be attributed to our hierarchical entailment mechanism via Einstein midpoint aggregation, which promotes stronger alignment between token-level representations in long contexts and the global token.

\section{Additional Results}

\subsection{Comparison with state-of-the-art hyperbolic VLMs}

\subsubsection{Experimental setup.}
We compare HyFL-CLIP with state-of-the-art hyperbolic VLMs, including MERU~\cite{desai2023hyperbolic}, HyCoCLIP~\cite{pal2024compositional}, and UNCHA~\cite{kim2026uncertainty}.
For each hyperbolic VLM, we further apply our hyperbolic fine-tuning framework to evaluate whether the proposed long-context adaptation is also beneficial for models already trained in hyperbolic space.
Since these models already operate in hyperbolic space, we omit the Euclidean-to-hyperbolic distillation loss when fine-tuning them.
We also include OpenCLIP~\cite{cherti2023reproducible} and HyFL-CLIP initialized from OpenCLIP for comparison.
All experiments are conducted using ViT-B models.

\subsubsection{Evaluation protocol.}
We evaluate zero-shot long-caption cross-modal retrieval on DOCCI~\cite{onoe2024docci}, DCI~\cite{urbanek2024picture}, Long-DCI~\cite{najdenkoska2024tulip}, and Urban-1k~\cite{zhang2024long}.
Each entry reports T2I / I2T Recall@1 (\%). The hyperbolic VLMs are trained on substantially smaller-scale pretraining data than OpenCLIP, using approximately $20.5$M image-text pairs compared with $2.3$B pairs for OpenCLIP.

\subsubsection{Experimental results.}
Tab.~\ref{tab:hvlm_ours} shows that applying our fine-tuning framework consistently improves existing hyperbolic VLMs across all datasets.
These results indicate that our method is effective not only for transferring Euclidean CLIP representations into hyperbolic space, but also for adapting existing hyperbolic VLMs to long-context scenarios.
Nevertheless, the fine-tuned hyperbolic VLMs still underperform HyFL-CLIP initialized from OpenCLIP, suggesting that leveraging a strong Euclidean VLM through efficient fine-tuning is both effective and computationally practical.

\begin{table}[t]
\centering
\caption{\textbf{Comparison with state-of-the-art hyperbolic VLMs on zero-shot long-caption cross-modal retrieval.}
Each entry reports T2I / I2T Recall@1 (
For hyperbolic VLMs, ``+ Ours'' denotes applying our hyperbolic fine-tuning framework without the Euclidean-to-hyperbolic distillation loss.}
\label{tab:hvlm_ours}

\scriptsize
\setlength{\tabcolsep}{3pt}
\renewcommand{\arraystretch}{0.95}

\begin{tabular}{lcccc}
\toprule
Model
& DOCCI~\cite{onoe2024docci}
& DCI~\cite{urbanek2024picture}
& Long-DCI~\cite{najdenkoska2024tulip}
& Urban-1k~\cite{zhang2024long} \\
\midrule
\multicolumn{5}{c}{\textit{State-of-the-art hyperbolic VLMs}} \\
\midrule
MERU~\cite{desai2023hyperbolic}
& 49.80 / 53.16
& 46.52 / 49.77
& 32.11 / 34.00
& 45.70 / 51.90 \\
MERU + Ours
& \textbf{58.10} / \textbf{61.71}
& \textbf{55.03} / \textbf{58.93}
& \textbf{41.08} / \textbf{44.75}
& \textbf{74.20} / \textbf{74.30} \\
\midrule
HyCoCLIP~\cite{pal2024compositional}
& 52.04 / 53.78
& 47.87 / 49.02
& 32.60 / 33.39
& 47.70 / 56.60 \\
HyCoCLIP + Ours
& \textbf{61.16} / \textbf{63.73}
& \textbf{58.88} / \textbf{60.68}
& \textbf{42.74} / \textbf{45.30}
& \textbf{75.30} / \textbf{76.70} \\
\midrule
UNCHA~\cite{kim2026uncertainty}
& 46.76 / 46.73
& 45.12 / 44.82
& 29.16 / 29.27
& 39.90 / 43.80 \\
UNCHA + Ours
& \textbf{63.57} / \textbf{66.92}
& \textbf{60.58} / \textbf{62.48}
& \textbf{44.23} / \textbf{47.03}
& \textbf{77.10} / \textbf{76.80} \\
\midrule
\multicolumn{5}{c}{\textit{Euclidean VLM with our hyperbolic fine-tuning}} \\
\midrule
OpenCLIP~\cite{cherti2023reproducible}
& 57.22 / 60.84
& 46.97 / 50.78
& 33.73 / 36.83
& 53.40 / 67.50 \\
\textbf{HyFL-CLIP (Ours)}
& \textbf{81.12} / \textbf{78.41}
& \textbf{71.79} / \textbf{71.54}
& \textbf{58.75} / \textbf{59.00}
& \textbf{91.10} / \textbf{91.80} \\
\bottomrule
\end{tabular}
\end{table}

\subsection{Full results of zero-shot long-caption cross-modal retrieval under caption perturbations.}
\begin{table*}[t]
\centering
\caption{\textbf{Full robustness comparison under caption perturbations.}
Each entry reports the average retrieval performance under a specific caption perturbation setting, including word dropout, sentence removal, order shuffling, and random subsampling. The average rows summarize performance across all evaluated datasets. Results marked with $*$ are evaluated using our own implementation, while results marked with $\dagger$ are obtained using checkpoints provided by the original authors.}
\label{tab:robustness_full}

\resizebox{\textwidth}{!}{%
\begin{tabular}{llcccccc}
\toprule
\multirow{2}{*}{Model}
& \multirow{2}{*}{Dataset}
& \multicolumn{2}{c}{Word Dropout}
& Sent. Removal
& Order Shuffling
& \multicolumn{2}{c}{Random Subsampling} \\
\cmidrule(lr){3-4}
\cmidrule(lr){5-5}
\cmidrule(lr){6-6}
\cmidrule(lr){7-8}
& & $p=0.3$ & $p=0.5$ & first & random & $n=2$ & $n=3$ \\
\midrule

\multirow{6}{*}{Long-CLIP$^\dagger$~\cite{zhang2024long}}
& ShareGPT4V~\cite{chen2024sharegpt4v} & 86.91 & 75.79 & 88.60 & 88.80 & 45.34 & 51.19 \\
& Urban1k~\cite{zhang2024long}          & 61.73 & 44.78 & 59.40 & 68.82 & 22.51 & 27.98 \\
& DOCCI~\cite{onoe2024docci}            & 59.56 & 59.66 & 51.03 & 59.72 & 25.41 & 30.17 \\
& DCI~\cite{urbanek2024picture}         & 49.48 & 38.74 & 45.77 & 53.09 & 23.58 & 28.70 \\
& DCI-Long~\cite{najdenkoska2024tulip}  & 34.84 & 25.45 & 32.23 & 38.53 & 15.65 & 18.43 \\
\rowcolor{gray!12}
& Average                               & 58.50 & 48.88 & 55.41 & 61.79 & 26.50 & 31.29 \\
\midrule

\multirow{6}{*}{HiMo-CLIP*~\cite{wu2025himo}}
& ShareGPT4V~\cite{chen2024sharegpt4v} & 94.80 & 86.27 & 95.60 & 96.93 & 48.14 & 57.08 \\
& Urban1k~\cite{zhang2024long}          & 73.47 & 54.98 & 68.95 & 81.16 & 24.77 & 32.74 \\
& DOCCI~\cite{onoe2024docci}            & 71.50 & 71.46 & 60.90 & 71.65 & 26.84 & 33.81 \\
& DCI~\cite{urbanek2024picture}         & 59.16 & 47.05 & 54.93 & 64.15 & 25.12 & 30.04 \\
& DCI-Long~\cite{najdenkoska2024tulip}  & 45.70 & 33.97 & 41.81 & 50.80 & 16.35 & 20.55 \\
\rowcolor{gray!12}
& Average                               & \underline{68.93} & \underline{58.75} & \underline{64.44} & 72.94 & \underline{28.24} & \underline{34.84} \\
\midrule

\multirow{6}{*}{FineLIP*~\cite{asokan2025finelip}}
& ShareGPT4V~\cite{chen2024sharegpt4v} & 93.74 & 85.48 & 94.80 & 96.23 & 45.34 & 53.42 \\
& Urban1k~\cite{zhang2024long}          & 73.45 & 54.19 & 68.20 & 82.57 & 24.70 & 31.96 \\
& DOCCI~\cite{onoe2024docci}            & 72.92 & 72.84 & 61.54 & 72.93 & 26.35 & 33.24 \\
& DCI~\cite{urbanek2024picture}         & 58.27 & 47.19 & 54.18 & 64.33 & 23.79 & 28.78 \\
& DCI-Long~\cite{najdenkoska2024tulip}  & 45.41 & 33.80 & 42.17 & 51.00 & 15.84 & 19.65 \\
\rowcolor{gray!12}
& Average                               & 68.76 & 58.70 & 64.18 & \underline{73.41} & 27.20 & 33.41 \\
\midrule

\multirow{6}{*}{\textbf{HyFL-CLIP (Ours)}}
& ShareGPT4V~\cite{chen2024sharegpt4v} & 95.40 & 95.44 & 96.45 & 97.37 & 56.52 & 63.59 \\
& Urban1k~\cite{zhang2024long}          & 77.60 & 77.14 & 75.45 & 86.08 & 27.64 & 36.25 \\
& DOCCI~\cite{onoe2024docci}            & 68.18 & 68.48 & 65.13 & 75.53 & 30.93 & 37.20 \\
& DCI~\cite{urbanek2024picture}         & 61.43 & 61.74 & 58.51 & 67.69 & 28.73 & 34.56 \\
& DCI-Long~\cite{najdenkoska2024tulip}  & 48.32 & 48.18 & 45.56 & 54.14 & 18.93 & 23.63 \\
\rowcolor{gray!20}
& Average                               & \textbf{70.19} & \textbf{70.20} & \textbf{68.22} & \textbf{76.16} & \textbf{32.55} & \textbf{39.05} \\

\bottomrule
\end{tabular}%
}
\end{table*}

\subsubsection{Experimental results.} 
Table~\ref{tab:robustness_full} shows the full zero-shot retrieval results under caption perturbations. HyFL-CLIP (Ours) achieves the best average performance across all perturbation settings, including word dropout, sentence removal, order shuffling, and random subsampling. These results demonstrate that our method is more robust to noisy, incomplete, and reordered long captions compared to existing long-caption VLMs.

\subsection{SDXL experiments with ViT-L}
\subsubsection{Experimental setup.} Following prior works~\cite{najdenkoska2024tulip, zhang2024long, asokan2025finelip}, we apply our HyFL-CLIP model to Stable Diffusion XL (SDXL)~\cite{podell2023sdxl} in a plug-and-play manner. In the original SDXL architecture, the text encoder consists of a CLIP-L encoder and an OpenCLIP bigG encoder. The prompts are encoded by both encoders, and the resulting text embeddings are concatenated to condition the image generation process. 

We replace the CLIP-L text encoder with our HyFL-CLIP-L. Since our model produces hyperbolic embeddings, these embeddings are mapped to the Euclidean space using the logarithmic map transport before concatenation. To extend the context length of the OpenCLIP bigG text encoder without retraining, we expand the positional embeddings via linear interpolation. The first 20 positional embeddings are preserved to maintain the well-trained short context structure. For the remaining positions, additional embeddings are inserted between consecutive positions through linear interpolation, since retraining this large encoder would be computationally expensive.

\subsubsection{Evaluation protocol.} We randomly sample $500$ prompts from COCO~\cite{lin2014microsoft}, DOCCI~\cite{onoe2024docci}, and Long-DCI~\cite{najdenkoska2024tulip}, and evaluate on the full DrawBench~\cite{saharia2022photorealistic} benchmark. DOCCI~\cite{onoe2024docci} and Long-DCI~\cite{najdenkoska2024tulip} are used to evaluate the model’s ability to generate images from long captions. Using the corresponding captions, we generate images with SDXL integrated with HyFL-CLIP. Following FineLIP~\cite{asokan2025finelip}, we evaluate image quality using the Fréchet Inception Distance (FID) and measure text–image alignment using CLIP similarity. Other baselines~\cite{zhang2024long, wu2025himo} are implemented using their respective ViT-L/14 backbone versions.

\label{sec:SDXL suppl}

\subsubsection{Experimental results.}  Fig.~\ref{fig:short_caption_SDXL}, Fig.~\ref{fig:DOCCI_SDXL}, Fig.~\ref{fig:Long-DCI_SDXL}, and Tab.~\ref{tab:t2i_results} present qualitative and quantitative comparisons between HyFL-CLIP (Ours) and other methods~\cite{wu2025himo, zhang2024long} when integrated into SDXL~\cite{podell2023sdxl}. The results demonstrate that our method integrates seamlessly with existing generation pipelines while producing high-quality text embeddings that remain well aligned with long captions.

\begin{table}[t]
\centering
\caption{\textbf{Frechet Inception Distance (FID) and CLIP~\cite{Radford2021LearningTV} similarity scores for text-to-image generation.} We randomly sample $500$ prompts from COCO~\cite{lin2014microsoft}, DOCCI~\cite{onoe2024docci}, and Long-DCI~\cite{najdenkoska2024tulip}, and evaluate on the full DrawBench~\cite{saharia2022photorealistic} benchmark. Our model achieves competitive or best performance in both image quality and text–image alignment. The downward arrow ($\downarrow$) indicates that lower values are better, while the upward arrow ($\uparrow$) indicates that higher values are better.
} 
\label{tab:t2i_results}
\resizebox{\linewidth}{!}{
\begin{tabular}{lccccccc}
\toprule
& \multicolumn{3}{c}{FID Score $\downarrow$} 
& \multicolumn{4}{c}{CLIP Similarity Score $\uparrow$} \\
\cmidrule(lr){2-4} \cmidrule(lr){5-8}
Model & {\scriptsize COCO} & {\scriptsize DOCCI} & {\scriptsize Long-DCI} & {\scriptsize COCO} & {\scriptsize DOCCI} & {\scriptsize Long-DCI} & {\scriptsize DrawBench} \\
\midrule
Long-CLIP~\cite{zhang2024long} & \textbf{26.64} & 26.03 & 30.62 & \underline{0.438} & \underline{0.446} & 0.435 & \textbf{0.427} \\
HiMo-CLIP~\cite{wu2025himo} & \underline{27.07} & \underline{25.92} & \underline{30.42} & 0.431 & 0.445 & \underline{0.437} & \underline{0.419} \\
\rowcolor{gray!15}
HyFL-CLIP (Ours) & 27.23 & \textbf{24.52} & \textbf{28.28} & \textbf{0.444} & \textbf{0.451} & \textbf{0.443} & \textbf{0.427} \\
\bottomrule
\end{tabular}
}
\end{table}

\begin{figure}[tb]
\centering

\begin{subfigure}[t]{\linewidth}
    \centering
    \includegraphics[width=\linewidth]{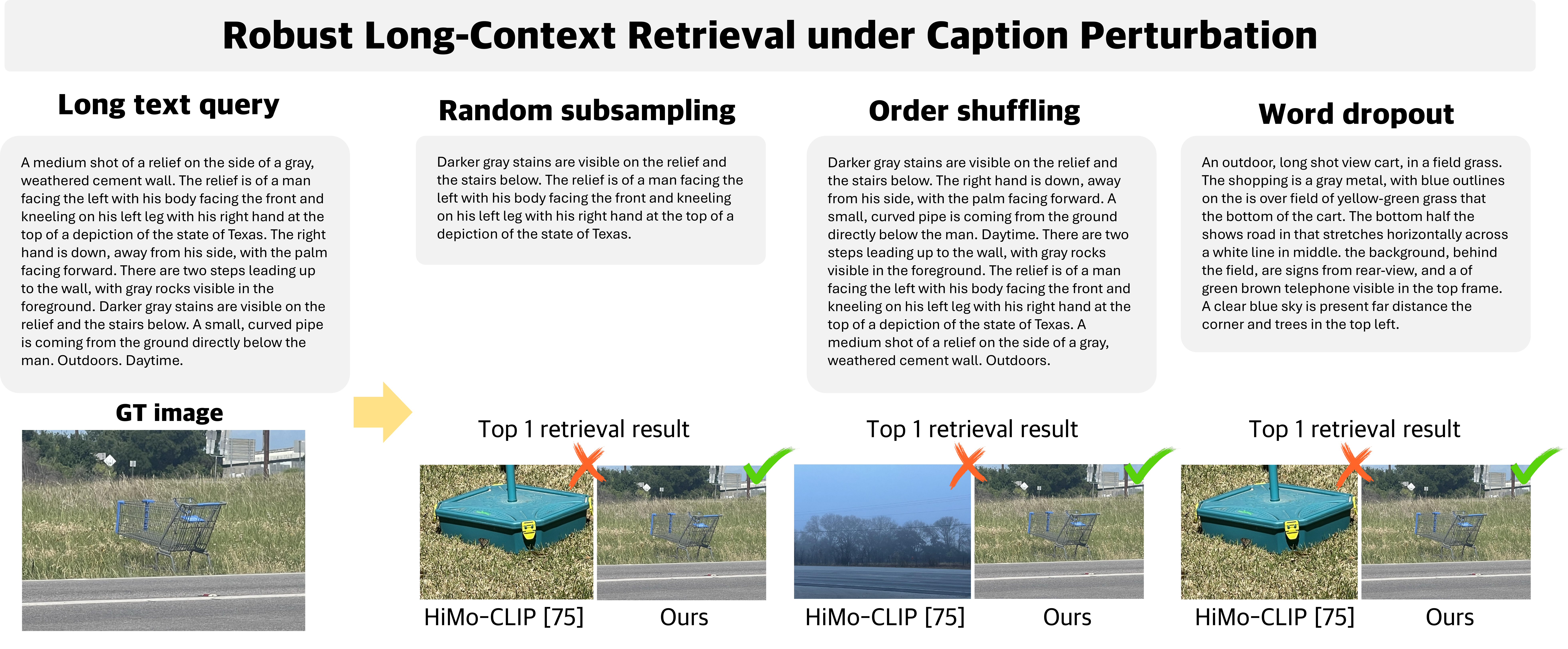}
    \caption{Example with a roadside cart scene. }
\end{subfigure}

\begin{subfigure}[t]{\linewidth}
    \centering
    \includegraphics[width=\linewidth]{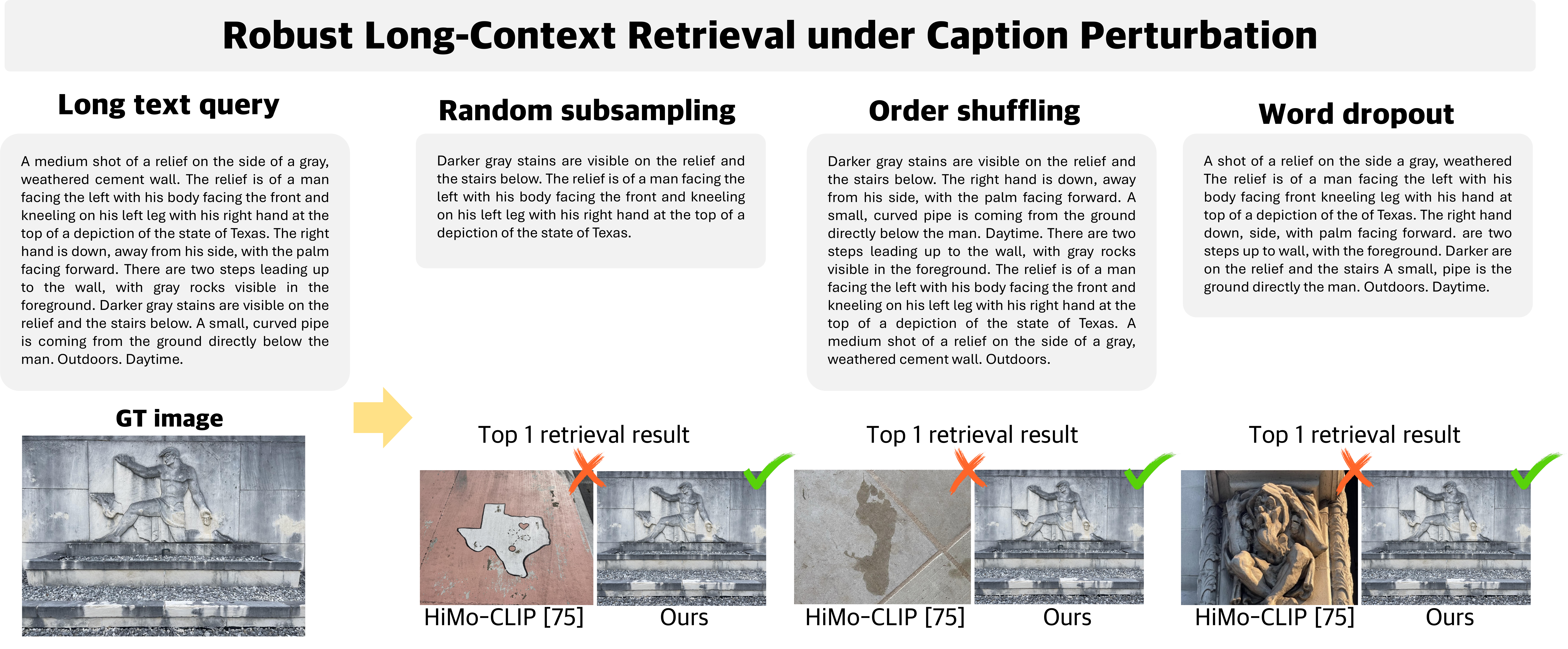}
    \caption{Example with a relief sculpture scene.}
\end{subfigure}

\caption{\textbf{Robust long-context retrieval under caption perturbations on Long-DCI.}
We compare retrieval results obtained with HiMo-CLIP and our HyFL-CLIP when captions are perturbed in different ways, including random subsampling, order shuffling, and word dropout.
For each query, we show the ground-truth (GT) image and the top-1 retrieval result from each method.
While HiMo-CLIP often retrieves incorrect images under caption perturbations, our method consistently retrieves the correct image, demonstrating improved robustness to long-context caption variations. 
}

\label{fig:robust_long_context_retreival_caption_peturb}

\end{figure}

\begin{figure}[tb]
  \includegraphics[width=\linewidth]{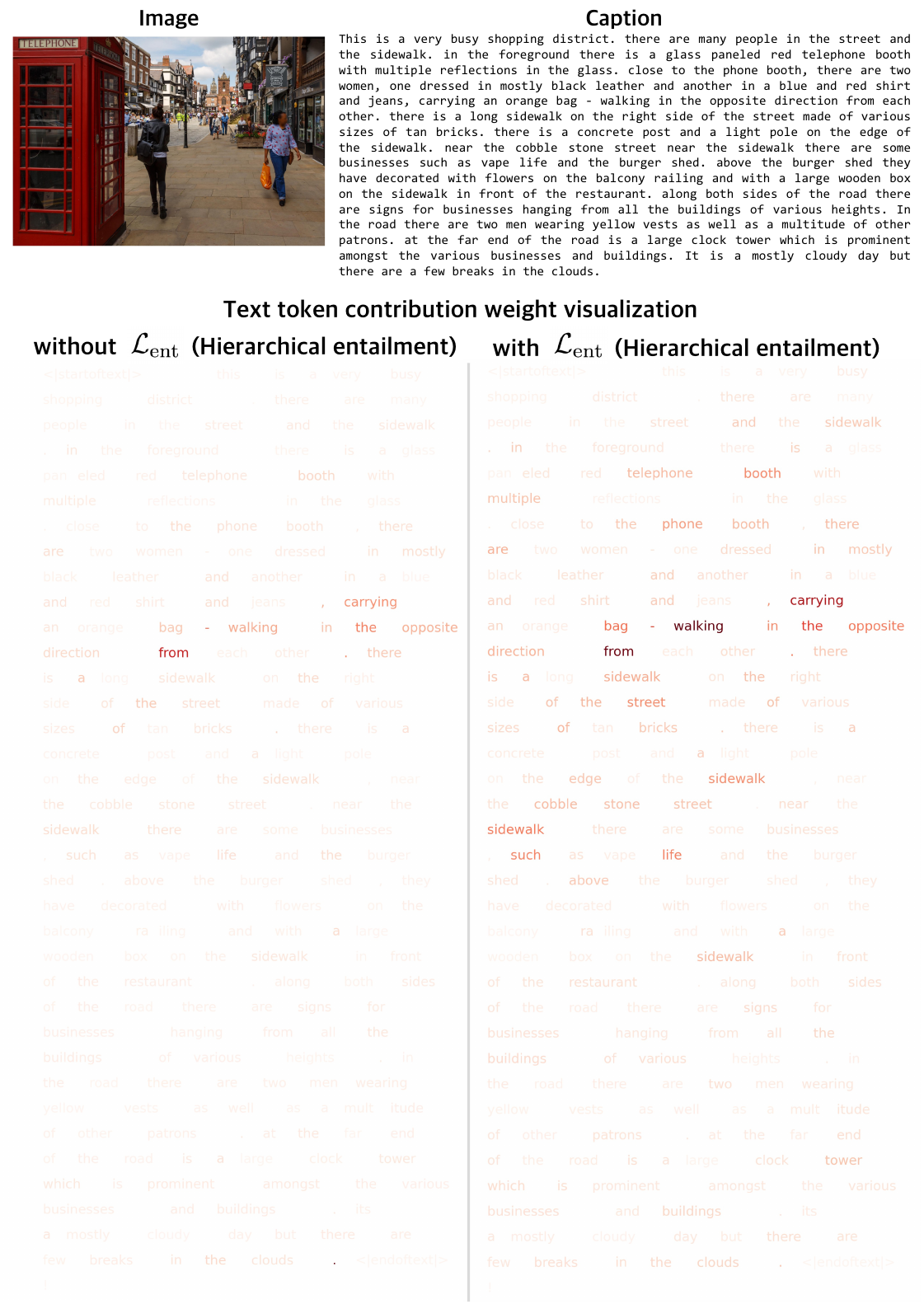}
\caption{\textbf{Visualization of text token contribution weights when trained with or without $\mathcal{L}_{ent}$ (hierarchical entailment).}
We visualize the contribution of individual text tokens to the merging process by measuring their similarity with the corresponding image representation. 
As shown in the figure, the model trained with $\mathcal{L}_{ent}$ highlights semantically important tokens more clearly, indicating that hierarchical entailment encourages the model to focus on informative words in the caption.}
   \label{fig:activation_viz1}
\end{figure}
\begin{figure}[tb]
  \includegraphics[width=\linewidth]{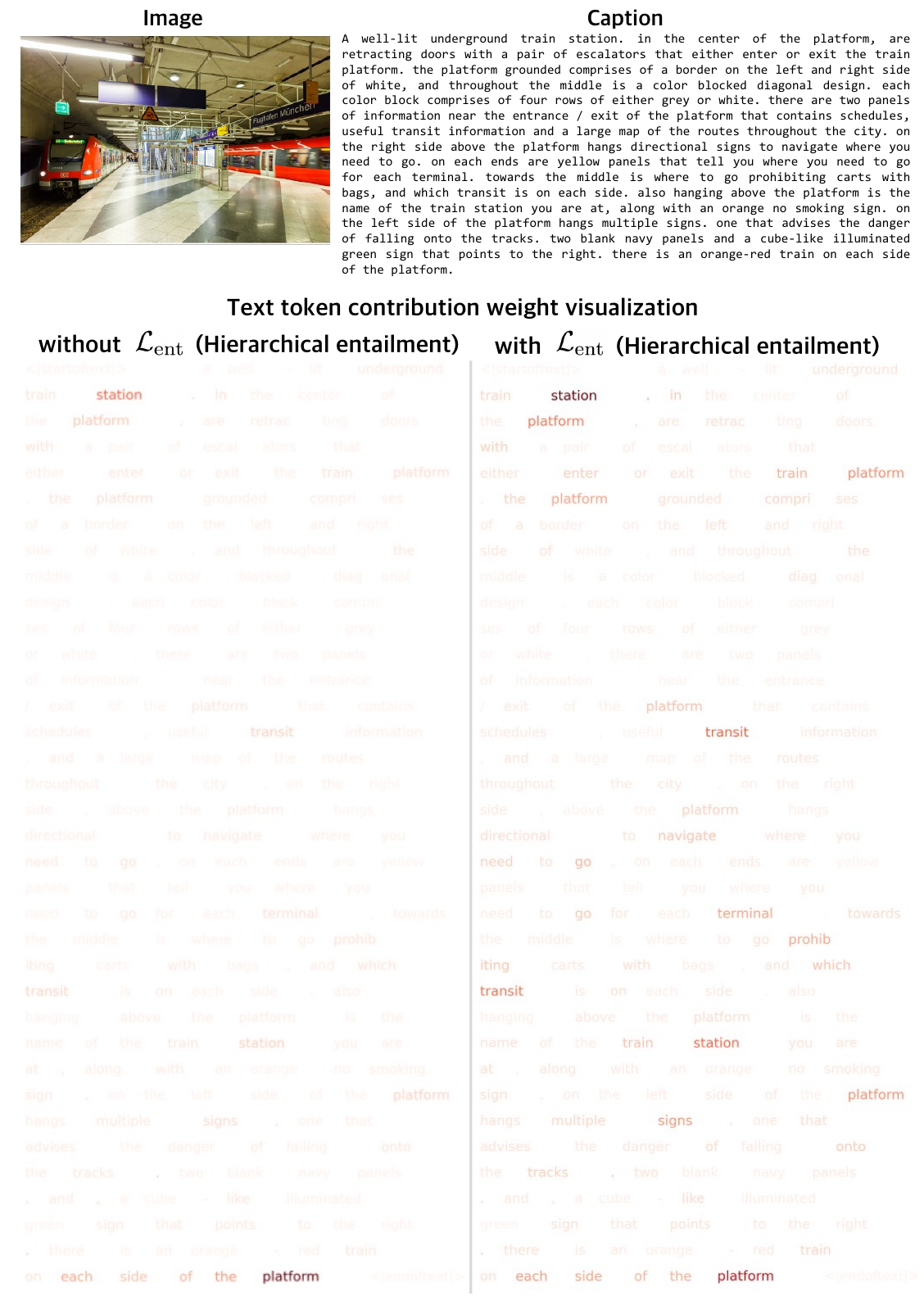}
 \caption{\textbf{Visualization of text token contribution weights when trained with or without $\mathcal{L}_{ent}$ (hierarchical entailment).}
We visualize the contribution of individual text tokens to the merging process by measuring their similarity with the corresponding image representation. 
As shown in the figure, the model trained with $\mathcal{L}_{ent}$ highlights semantically important tokens more clearly, indicating that hierarchical entailment encourages the model to focus on informative words in the caption.}
   \label{fig:activation_viz2}
\end{figure}

\begin{figure}[tb]
  \includegraphics[width=\linewidth]{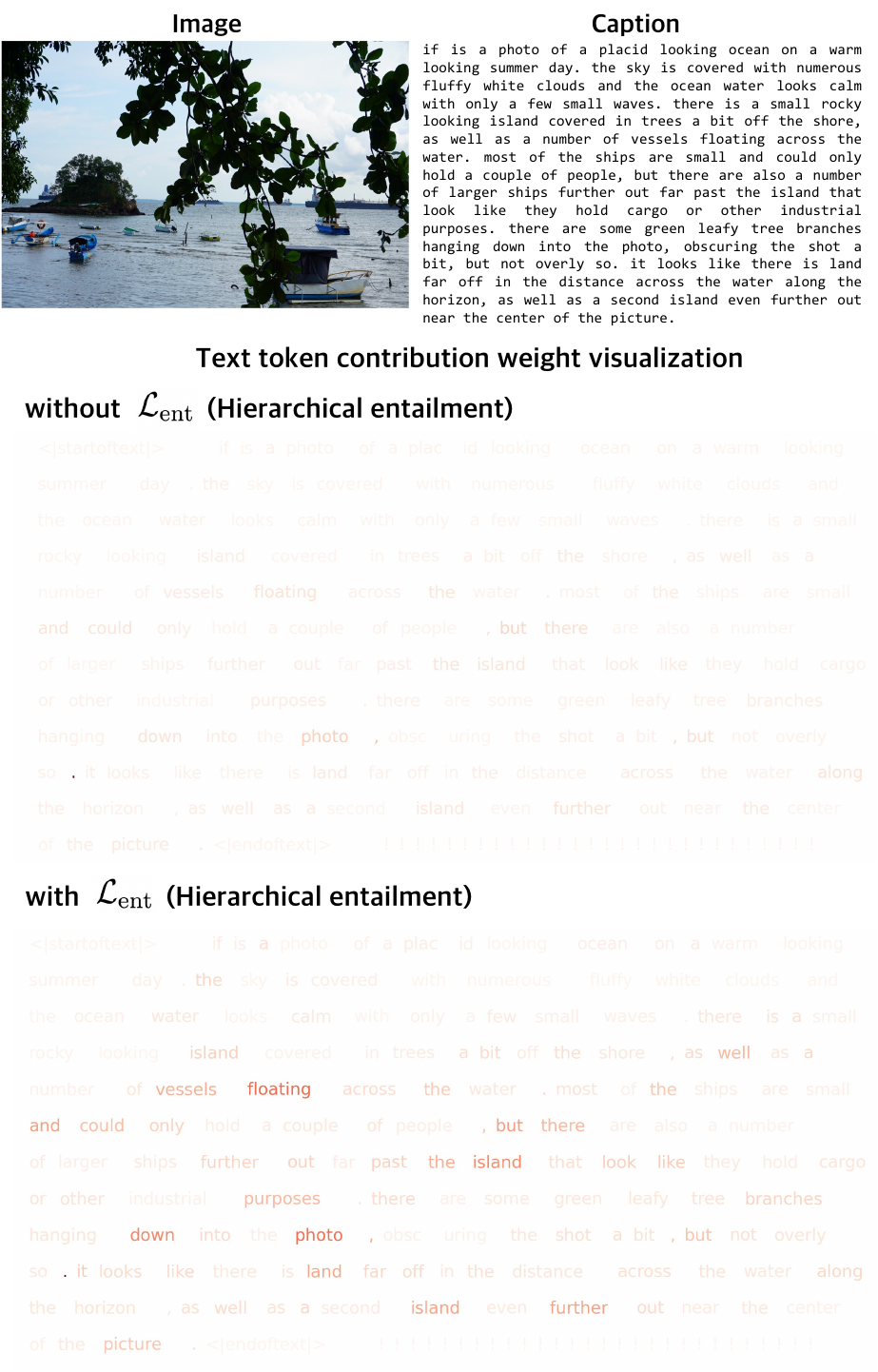}
 \caption{\textbf{Visualization of text token contribution weights when trained with or without $\mathcal{L}_{ent}$ (hierarchical entailment).}
We visualize the contribution of individual text tokens to the merging process by measuring their similarity with the corresponding image representation. 
As shown in the figure, the model trained with $\mathcal{L}_{ent}$ highlights semantically important tokens more clearly, indicating that hierarchical entailment encourages the model to focus on informative words in the caption.}
   \label{fig:activation_viz3}
\end{figure}

\begin{figure}[tb]
  \includegraphics[width=\linewidth]{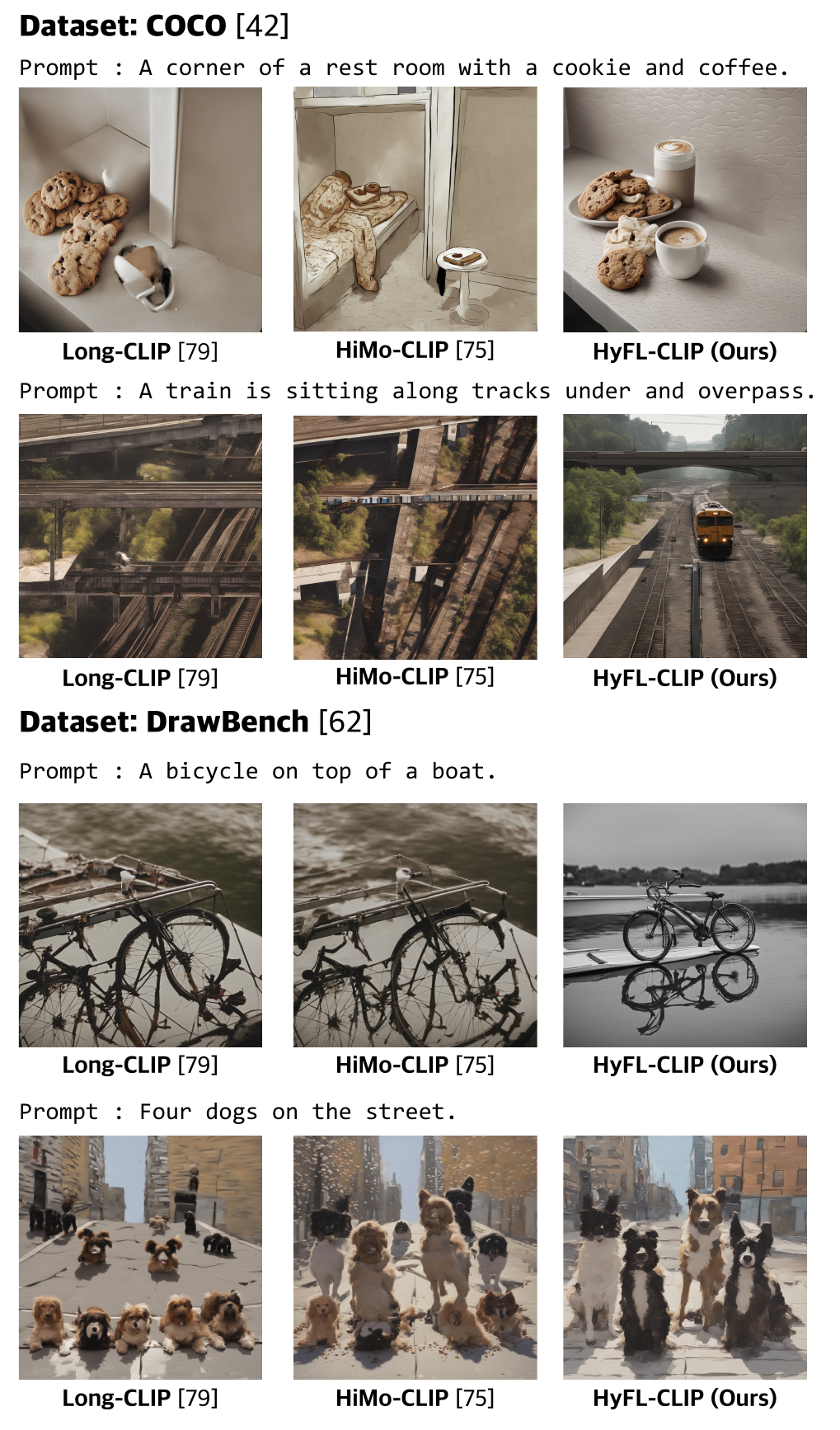}
 \caption{\textbf{Comparison of images generated from COCO~\cite{lin2014microsoft} captions using HyFL-CLIP (Ours) integrated SDXL and baselines.} Images generated with our model preserve finer visual details and exhibit higher fidelity to the given captions compared to the baseline methods. This demonstrates that HyFL-CLIP provides more precise semantic guidance for text-to-image generation.}

   \label{fig:short_caption_SDXL}
\end{figure}

\begin{figure}[tb]
  \includegraphics[width=\linewidth]{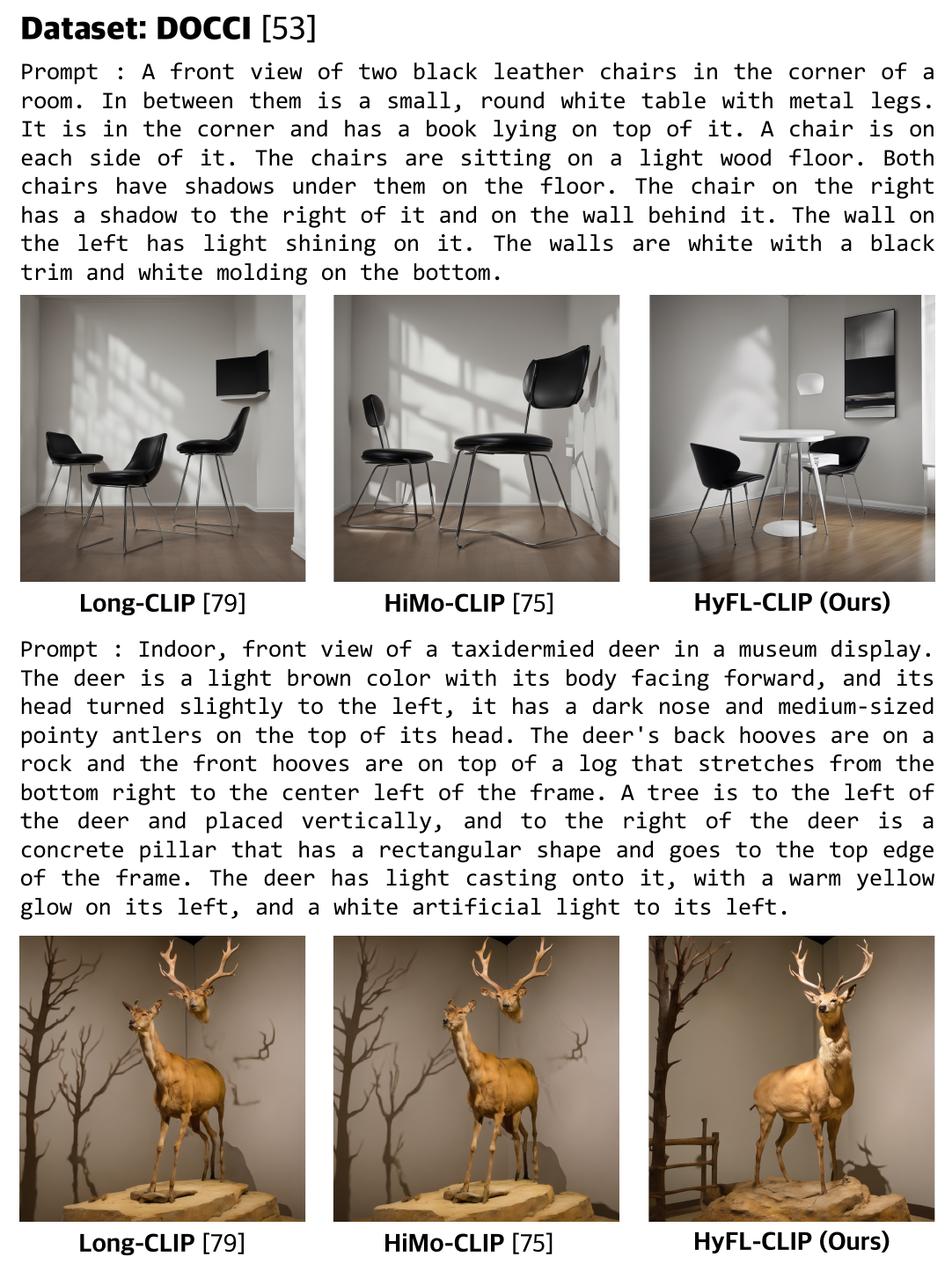}
 \caption{\textbf{Comparison of images generated from DOCCI~\cite{onoe2024docci} captions using HyFL-CLIP (Ours) integrated SDXL and baselines.} Images generated with our model preserve finer visual details and exhibit higher fidelity to the given captions compared to the baseline methods. This demonstrates that HyFL-CLIP provides more precise semantic guidance for text-to-image generation.}

   \label{fig:DOCCI_SDXL}
\end{figure}

\begin{figure}[tb]
  \includegraphics[width=\linewidth]{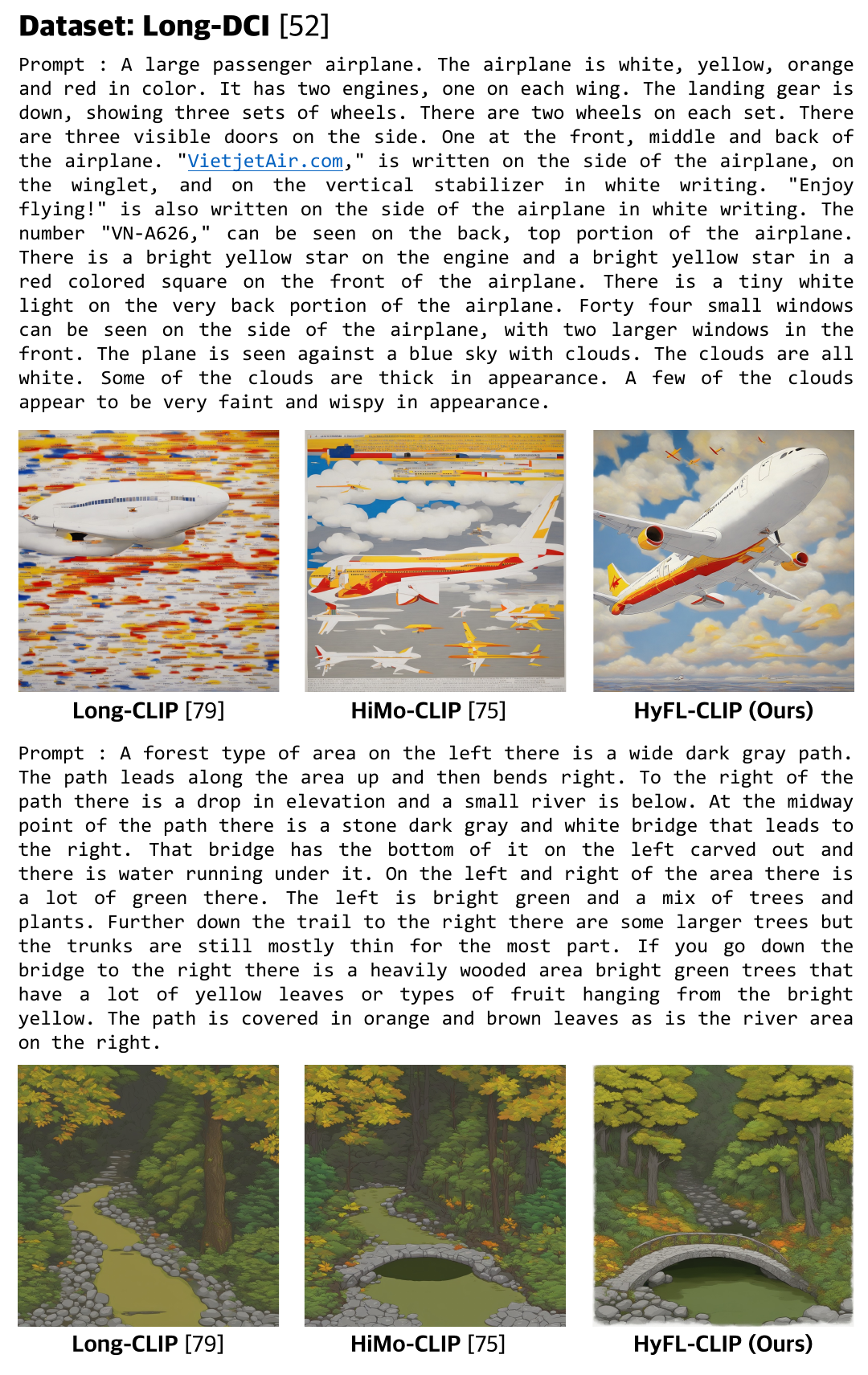}
 \caption{\textbf{Comparison of images generated from Long-DCI~\cite{najdenkoska2024tulip} captions using HyFL-CLIP (Ours) integrated SDXL and baselines.} Images generated with our model preserve finer visual details and exhibit higher fidelity to the given captions compared to the baseline methods. This demonstrates that HyFL-CLIP provides more precise semantic guidance for text-to-image generation.}

   \label{fig:Long-DCI_SDXL}
\end{figure}

\subsection{Additional examples of text-token contribution weight visualization}

We visualize the contribution of each text token to its paired target image to analyze how individual tokens influence image–text alignment. Specifically, given a long caption, we compute the similarity between each caption token and the paired image and use these similarities to weight the token embeddings when aggregating them into a single representation during training. The resulting weights quantify the contribution of each token to the aggregated representation, and we visualize these weights to analyze the contribution of individual tokens.

We visualize these contribution weights for each token in Fig.~\ref{fig:activation_viz1}, Fig.~\ref{fig:activation_viz2}, and Fig.~\ref{fig:activation_viz3}. In each figure, the image and the full caption are displayed at the top, while the bottom-left and bottom-right panels correspond to the models trained without and with the entailment loss ($\mathcal{L}_{ent}$), respectively. Darker red indicates a larger contribution weight. Without $\mathcal{L}_{ent}$, the contribution weights tend to be more uniformly distributed across tokens, whereas the model trained with $\mathcal{L}_{ent}$ assigns larger weights to semantically informative tokens that are strongly related to the visual content.

This behavior arises because our hierarchical entailment formulation aggregates token representations with different weights depending on how strongly each token contributes to the image–text alignment. As a result, the model emphasizes semantically relevant tokens when forming the final caption representation, leading to better alignment with the visual content. We further visualize token activations from HiMo-CLIP~\cite{wu2025himo} using the same procedure (see Fig.~\ref{fig:activation_comparison_viz1}, Fig.~\ref{fig:activation_comparison_viz2}, and Fig.~\ref{fig:activation_comparison_viz3}), and observe that our model activates a richer set of contextually relevant tokens compared to HiMo-CLIP~\cite{wu2025himo}, indicating that our formulation captures more informative semantic cues from long captions.

\subsection{More examples on long-text retrieval under caption perturbation}
We provide additional qualitative examples corresponding to Fig.~1 to further illustrate the robustness of our model under caption perturbations. For each long text query paired with a ground-truth (GT) image, we show the full captions after applying three perturbation strategies: random subsampling, order shuffling, and word dropout. The modified captions are presented at the top of the figure to highlight how the textual input changes under each perturbation. At the bottom, we display the top-1 retrieved images produced by HiMo-CLIP~\cite{wu2025himo} and our method using the perturbed captions as queries.

In Fig.~\ref{fig:robust_long_context_retreival_caption_peturb}(a), the perturbed queries produced by subsampling and shuffling still clearly describe a statue of a man scene. However, HiMo-CLIP~\cite{wu2025himo} retrieves unrelated asphalt images instead of the correct relief sculpture scene. Even in the word-dropout case, the retrieved result fails to reflect the pose of the man described in the caption.
A similar pattern appears in Fig.~\ref{fig:robust_long_context_retreival_caption_peturb}(b). Although the key object ``shopping cart” remains present in all perturbed captions, HiMo-CLIP~\cite{wu2025himo} consistently retrieves irrelevant images as the top result. In contrast, our method continues to retrieve the correct scene across all perturbation types.
These examples demonstrate that our model is significantly more robust to caption perturbations than existing methods. For completeness, the full captions used in the experiments corresponding to Fig.~1 are provided in Tab.~\ref{tab:full_caption_fig1}.

\begin{figure}[tb]
  \includegraphics[width=\linewidth]{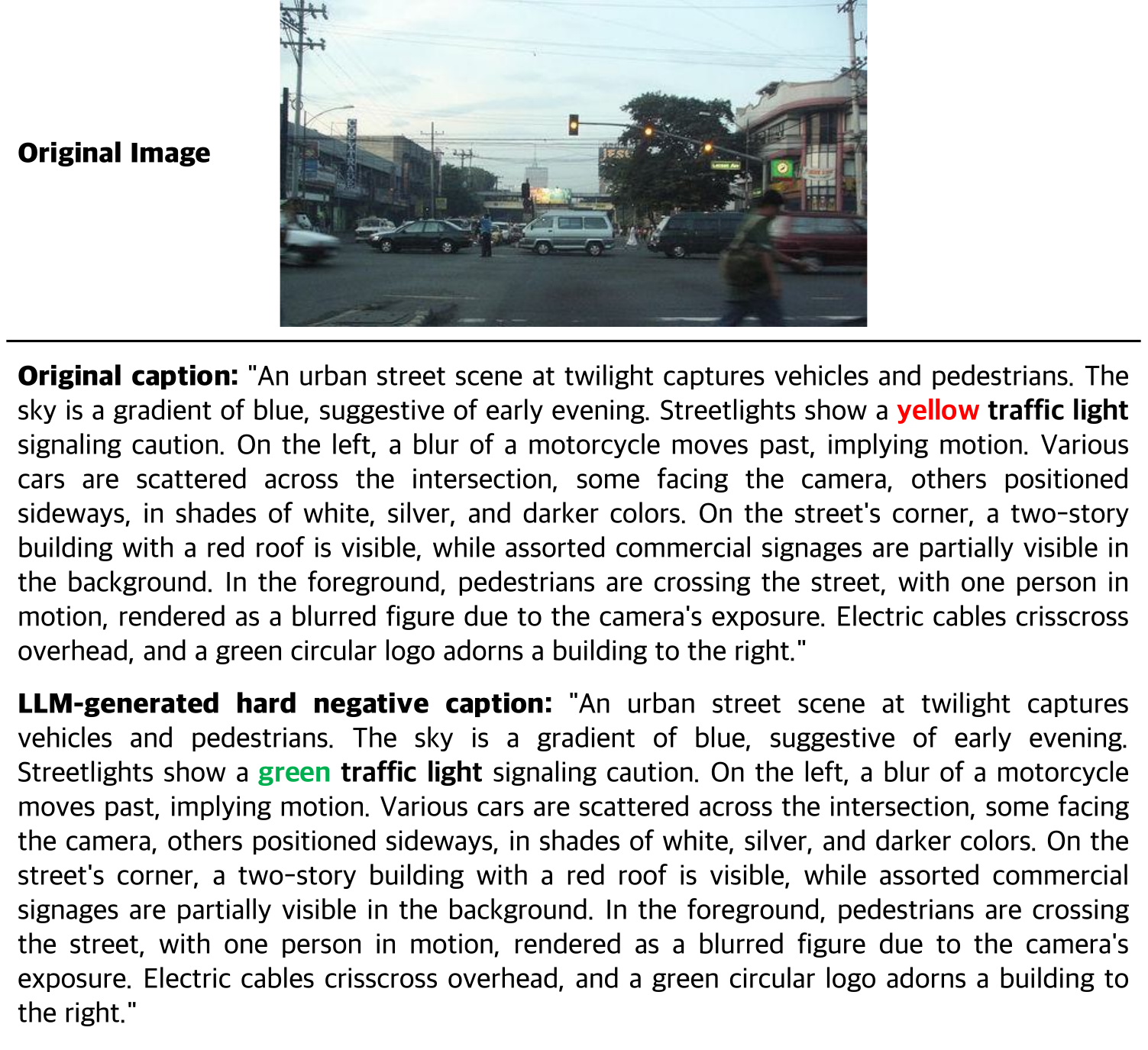}
 \caption{\textbf{Example of an LLM-generated hard-negative caption from Urban-1k~\cite{zhang2024long}.} A single-word perturbation (\textit{yellow} → \textit{green}) produces a highly subtle yet semantically incorrect caption, illustrating the difficulty of distinguishing such hard negatives.}
   \label{fig:hard_neg}
\end{figure}

\begin{figure}[tb]
  \includegraphics[width=\linewidth]{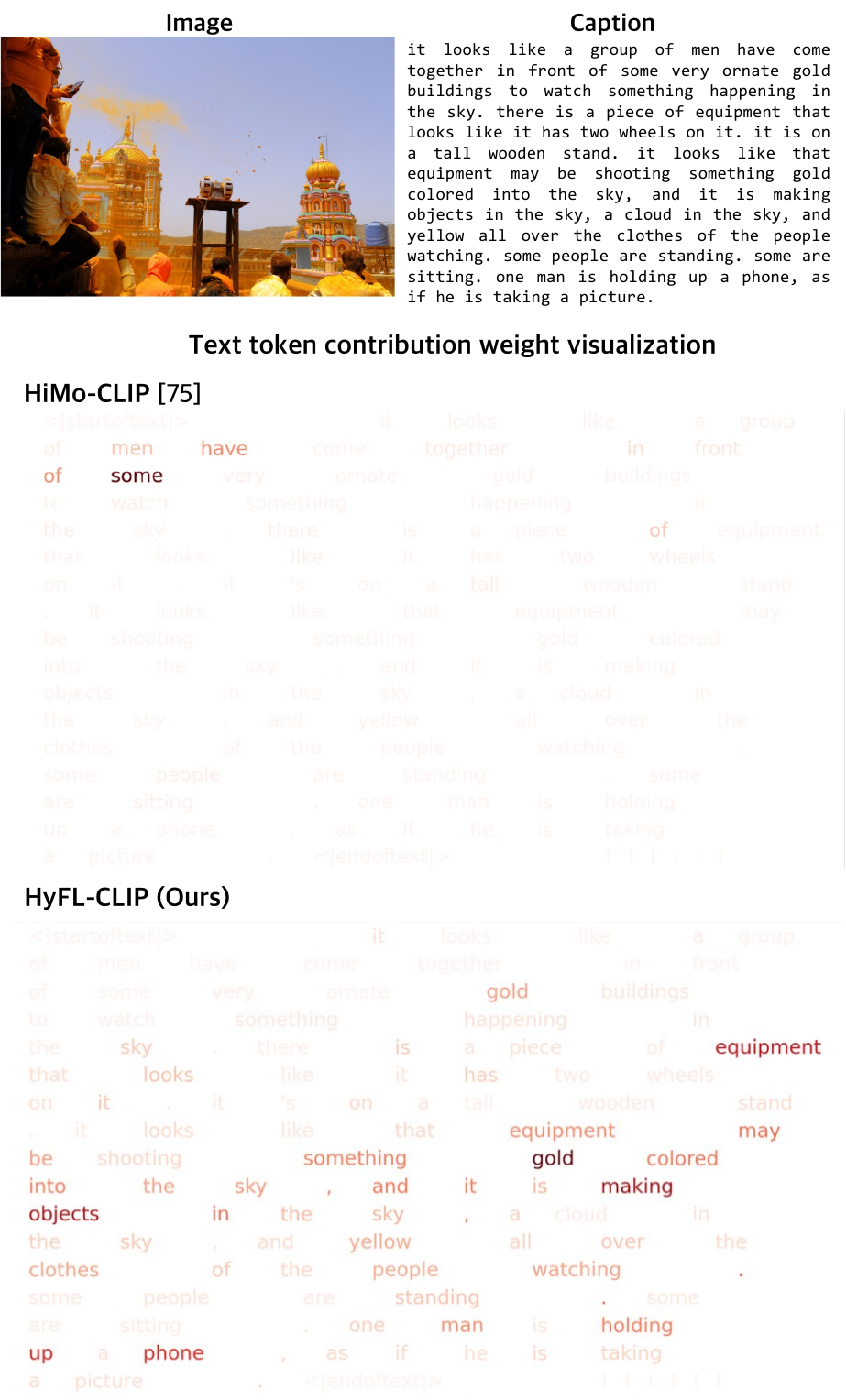}
 \caption{\textbf{Token contribution comparison between HiMo-CLIP~\cite{wu2025himo} and HyFL-CLIP (Ours).} For the same image-caption pair, HyFL-CLIP assigns higher weights to semantically meaningful tokens that correspond to visual elements in the image.}
   \label{fig:activation_comparison_viz1}
   
\end{figure}
\begin{figure}[tb]
  \includegraphics[width=\linewidth]{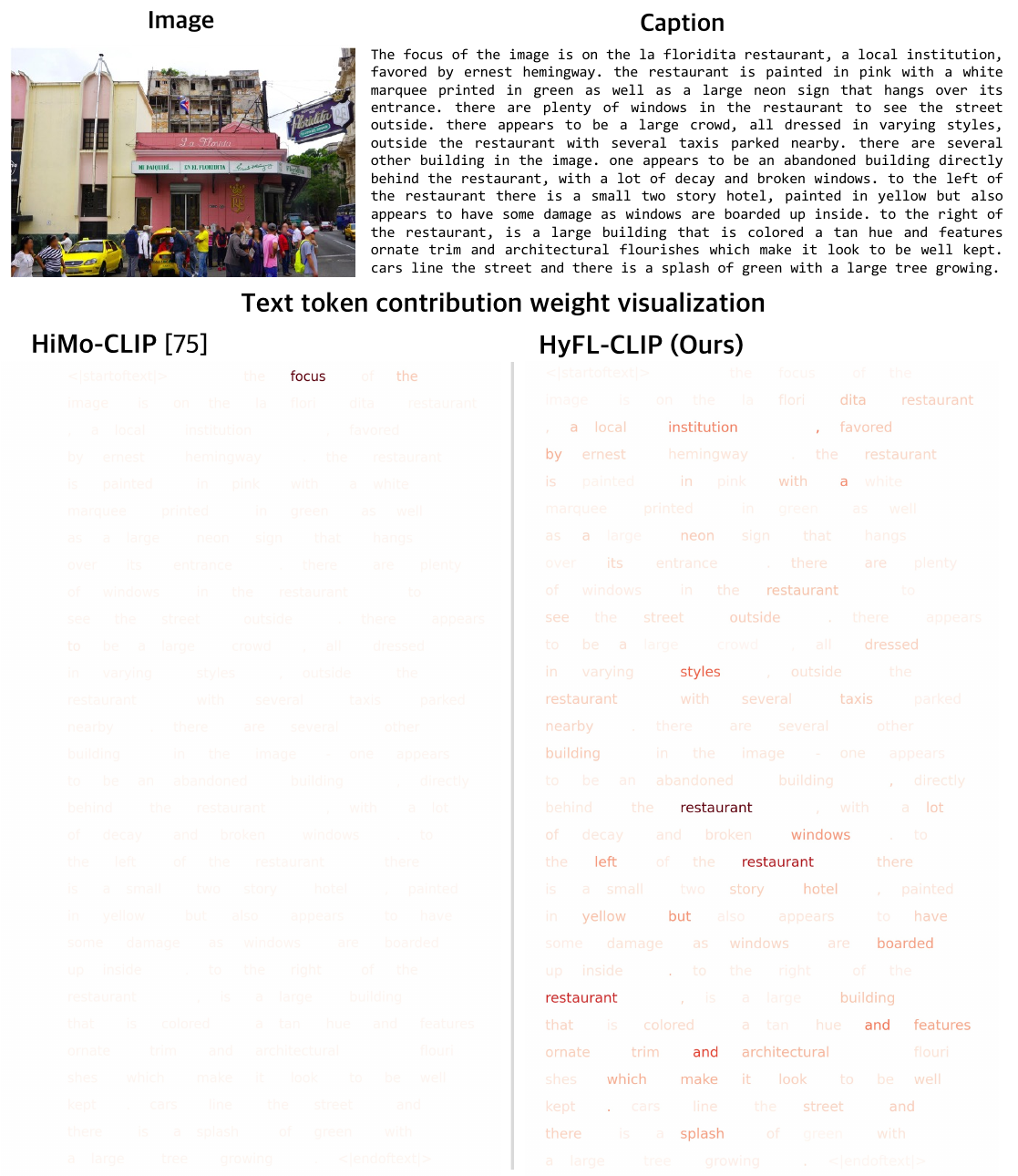}
\caption{\textbf{Token contribution comparison between HiMo-CLIP~\cite{wu2025himo} and HyFL-CLIP (Ours).} For the same image-caption pair, HyFL-CLIP assigns higher weights to semantically meaningful tokens that correspond to visual elements in the image.}
   \label{fig:activation_comparison_viz2}
\end{figure}
\begin{figure}[tb]
  \includegraphics[width=\linewidth]{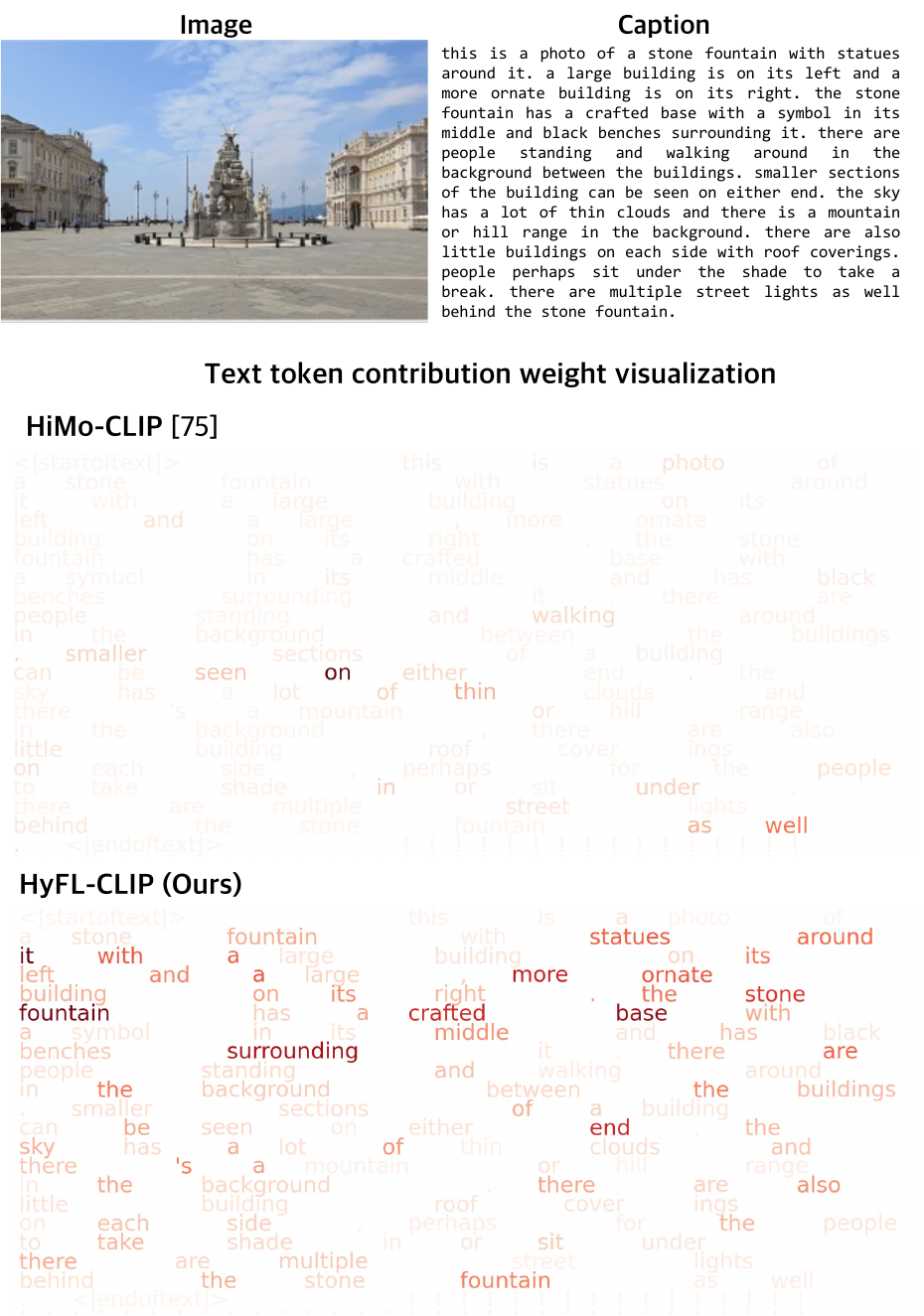}
\caption{\textbf{Token contribution comparison between HiMo-CLIP~\cite{wu2025himo} and HyFL-CLIP (Ours).} For the same image-caption pair, HyFL-CLIP assigns higher weights to semantically meaningful tokens that correspond to visual elements in the image.}
   \label{fig:activation_comparison_viz3}
\end{figure}
\

\begin{table*}[t]
\centering
\small
\setlength{\tabcolsep}{6pt}
\renewcommand{\arraystretch}{1.2}

\caption{\textbf{Examples of caption perturbations applied to a long text query.}
We show the full caption in Fig.1 and its perturbed variants generated by random subsampling, order shuffling, and word dropout. These perturbations preserve partial semantic information while altering the structure or completeness of the caption.}
\label{tab:full_caption_fig1}

\begin{tabularx}{\textwidth}{lX}
\toprule

\textbf{Type} & \textbf{Caption} \\

\midrule

\textbf{Long text query} &
The image shows an urban street intersection with vehicular and pedestrian activity. 
In the foreground, there is a pedestrian crossing the street marked with white zebra lines and a black car in the middle of the crosswalk. 
The traffic light for pedestrians is visible, displaying a red hand signal indicating a ``Do Not Walk'' command. 
A man wearing a white jacket and dark pants is walking away from the camera while carrying a black bag in his left hand. 
To the left, another pedestrian wearing a white and red outfit is crossing the street. 
Buildings with varying facades line the street, and clear blue skies with scattered clouds appear above. \\

\midrule

\textbf{Random subsampling} &
To the left, another pedestrian wearing a white and red outfit is crossing the street. 
The traffic light for pedestrians is visible, displaying a red hand signal indicating a ``Do Not Walk'' command. 
The image shows an urban street intersection with vehicular and pedestrian activity. \\

\midrule

\textbf{Order shuffling} &
Buildings with varying facades line the street, and clear blue skies with scattered clouds appear above. 
The traffic light for pedestrians is visible, displaying a red hand signal indicating a ``Do Not Walk'' command. 
The image shows an urban street intersection with vehicular and pedestrian activity. 
To the left, another pedestrian wearing a white and red outfit is crossing the street. 
In the foreground, there is a pedestrian crossing the street marked with white zebra lines and a black car in the middle of the crosswalk. 
A man wearing a white jacket and dark pants is walking away from the camera while carrying a black bag in his left hand. \\

\midrule

\textbf{Word dropout} &
The image shows an urban street intersection with vehicular and pedestrian activity. 
In the foreground, a crossing the marked with white zebra lines and a black car middle the. 
The traffic light for is visible, displaying a red hand signal indicating a ``Do Not Walk'' command. 
A man dark is walking the camera, black bag in his hand. 
Another pedestrian appears wearing a white and red the street. 
Buildings facades line the street, clear blue skies with scattered clouds. \\

\bottomrule
\end{tabularx}

\end{table*}

{   \clearpage
    \small
    \bibliographystyle{splncs04}
    \bibliography{main}
}

\end{document}